\documentclass[10pt,final,journal]{IEEEtran}
\usepackage{cite}
 \usepackage{graphics}
 \usepackage{graphicx}
\usepackage{hyperref}

\usepackage{algorithm}
\usepackage{algorithmic}

\usepackage{multirow}
\usepackage{array}
\usepackage{amsfonts}
\usepackage{amsmath}
\usepackage{mathrsfs}
\usepackage{color}
\usepackage{graphicx}
 \usepackage{subfigure}
\newcommand{\algorithmicinput}{\textbf{Input:}}
\newcommand{\algorithmicoutput}{\textbf{Output:}}
\newcommand{\algorithmicinitialization}{\textbf{Initialize:}}
\newcommand{\algorithmicsteps}{\textbf{Iterative:}}
\newcommand{\algorithmicNewUntil}{\textbf{Until}}
\newcommand{\INPUT}{\item[\algorithmicinput]}
  \newcommand{\OUTPUT}{\item[\algorithmicoutput]}
  \newcommand{\Initialization}{\item[\algorithmicinitialization]}
  \newcommand{\STEPS}{\item[\algorithmicsteps]}
    \newcommand{\NewUNTIL}{\item[\algorithmicNewUntil]}

\begin{document}

\title{Enhancing Person Re-identification in a Self-trained Subspace}
\author{Xun~Yang,
        Meng~Wang, %~\IEEEmembership{Member,~IEEE,}
        Richang Hong,
      Qi Tian, %~\IEEEmembership{Fellow,~IEEE,}
      Yong Rui %~\IEEEmembership{Fellow,~IEEE,}
\thanks{Xun Yang, Meng Wang, and Richang Hong are with the School
of Computer Science and Information Engineering, Hefei University of Technology, Hefei, China (\{hfutyangxun, eric.mengwang, hongrc.hfut\}@gmail.com).
 
 Qi Tian is with the Department of Computer Science, University of Texas at San Antonio (qitian@cs.utsa.edu).
 
  Yong Rui is with Lenovo (yongrui@lenovo.com). 
}}
\maketitle
\begin{abstract}
Despite the promising progress made in recent years, person re-identification (re-ID) remains a challenging task due to the complex variations in human appearances from different camera views. For this challenging problem, a large variety of algorithms have been developed in the fully-supervised setting, requiring access to a large amount of labeled training data. However, the main bottleneck for fully-supervised re-ID is the limited availability of labeled training samples. To address this problem, in this paper, we propose a self-trained subspace learning paradigm for person re-ID which effectively utilizes both labeled and unlabeled data to learn a discriminative subspace where person images across disjoint camera views can be easily matched.
The proposed approach first constructs pseudo pairwise relationships among unlabeled persons using the k-nearest neighbors algorithm. Then, with the pseudo pairwise relationships, the unlabeled samples can be easily combined with the labeled samples to learn a discriminative projection by solving an eigenvalue problem. In addition, we refine the pseudo pairwise relationships iteratively, which further improves the learning performance. A multi-kernel embedding strategy is also incorporated into the proposed approach to cope with the non-linearity in person's appearance and explore the complementation of multiple kernels. In this way, the performance of person re-ID can be greatly enhanced when training data are insufficient. Experimental results on six widely-used datasets demonstrate the effectiveness of our approach and its performance can be comparable to the reported results of most state-of-the-art fully-supervised methods while using much fewer labeled data. 

Our code is available: \\
{\color{blue} {\url{https://github.com/Xun-Yang/ReID_slef-training_TOMM2017}}}.

\end{abstract}
%\IEEEpeerreviewmaketitle

\section{Introduction}
\label{Section1}
Person Re-identification (re-ID) \cite{zheng2016Survey,LOMOXQDA,GaussianReIDDescriptor,ZhengLiangReIDBenchmark,Zheng2016,ZhengWeishi_AsymmetricDistance} aims to recognize an individual across spatially disjoint cameras. It has attracted much attention in recent years for its great potential in surveillance applications such as crowded scenes anomaly detection \cite{RN5} and multi-cameras pedestrian tracking \cite{RN4}. Although a large number of approaches have been proposed for re-ID, it remains a challenging problem since a person's appearance often undergoes dramatic changes across camera views due to changes in view angle, body pose, illumination and background clutter.

The fundamental re-ID problem is to compare a person of interest seen in a probe camera view to a gallery of candidates captured from a camera that does not overlap
with the probe one. If a true match to the probe exists in the gallery, it should have a high matching/similarity score, or rank, compared to incorrect candidates.
Generally, there are two basic problems: (1) feature representation and (2) metric learning. An effective feature representation \cite{GaussianReIDDescriptor,LOMOXQDA,Invariant_Color_FeaturesReID,WhosDescriptors} is critical for person re-ID, which should be robust to complex variations in human appearances from different camera views.
Several approaches have been investigated to design a feature descriptor directly based on low-level visual features.
More efforts \cite{KLFDA,zhang2016learning,KISSme,LADFReID,ZhengWeishi_AsymmetricDistance,ChenDaPeng_similarityLearning,MLAPG} have been made following the second direction to learn an optimal distance or similarity function to rank the potential matches based on their relevance. Some of them directly learn a Mahalanobis distance function parameterized by a positive semi-definite (PSD) matrix to separate positive person image pairs from negative pairs. Some others formulate re-ID as a subspace learning problem by learning a low-dimensional projection.
This work follows the second approach, aiming to learn a discriminative projection to map person images from disjoint camera views into a common subspace.

\begin{figure*}[tbp]
  \begin{center}
  \includegraphics[width=6in]{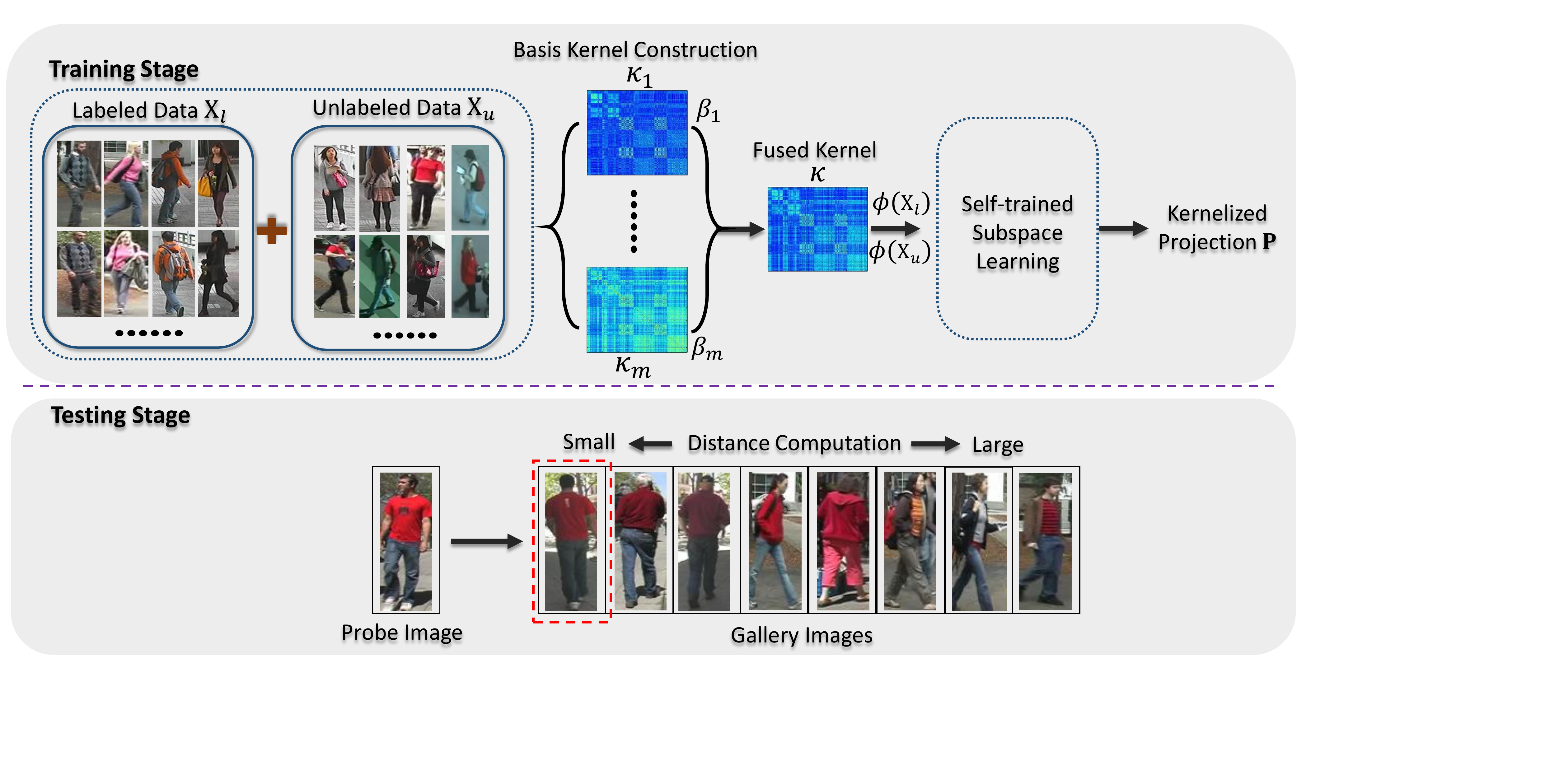}
  \end{center}
 %  \vspace{-0.1500in}
  \caption{Schematic illustration of the proposed person re-identification approach.}
 % \vspace{-0.00in}
  \label{figure1}
\end{figure*}

Despite the promising efforts made by many researchers, most existing methods are developed in the fully-supervised setting, requiring access to a large amount of labeled training image pairs. It is impractical to expect the availability of large quantities of labeled data because labeling data is very costly. The main bottleneck for fully-supervised re-ID is the limited availability of labeled training samples.
When only a small number of labeled data are available, supervised methods tend to learn a distance function that is over-fitted to the labeled data, which makes the learned distance function cannot generalize well to the test set. It's of great interest to design a solution that can utilize abundant unlabeled data. 
Although some semi-supervised re-ID approaches \cite{liu2014semi,IterativeLaplacian,SSMFL2013} have been proposed, their performances are far from satisfactory.

In this work, we design a self-trained subspace learning approach for person re-ID which effectively utilizes both labeled and unlabeled data to learn a discriminative subspace where person images across disjoint camera views can be easily matched. The classic self-training strategy \cite{zhu2009introduction} is exploited in this work. We first learn an initial projection matrix using the available labeled data only. Using this initial projection, all unlabeled person images are projected into a low-dimensional subspace, where the low-dimensional representation has higher discriminative power than the original features. Then, to utilize the unlabeled data, we construct pseudo pairwise relationships among the unlabeled persons using k-nearest neighbors (KNN) algorithm in this low-dimensional subspace. 
The pseudo pairwise relationships are encoded into a graph Laplacian regularization term which is further combined with a fully-supervised discriminative term to learn a new projection. 
Given the newly learned projection, we refine the pseudo pairwise relationships and relearned the discriminative projection with these updated pseudo pairwise relationships. This process is iterated until the pseudo pairwise relationships remain unchanged.
In this way, the discriminant power of the learned subspace will be enhanced.
 Besides, a multi-kernel embedding strategy is incorporated into the proposed approach to cope with the non-linearity in person's appearance and explore the complementation of multiple kernels. The final person matching can be performed very efficiently by computing the Euclidean distance between a probe image and a gallery image in the self-trained subspace. A schematic illustration of the proposed approach is shown in Fig. \ref{figure1}.

Our main contributions are summarized as follows:

(1) We propose an effective self-trained subspace learning framework for person re-ID which is able to utilize both labeled and unlabeled person images effectively. An iterative learning strategy is included to update the pseudo pairwise relationships among unlabeled persons.

(2) We introduce a multiple kernel embedding technique into the self-trained subspace learning framework, which explores the complementary information shared by multiple kernels and handles the non-linearity in person's appearance effectively.

(3) We conduct empirical studies on widely-used person re-ID datasets. Experimental results demonstrate that the proposed method is able to achieve a performance on par with the reported results of most state-of-the-art fully-supervised methods while using much fewer labeled person samples.

\section{Related Work}
\label{Section2}
\subsection{Person Re-ID}
During the past decades, many person re-ID algorithms \cite{zheng2017personPRW,LOMOXQDA,GaussianReIDDescriptor,VIPeR,Zheng2015,KISSme,KLFDA,LADFReID,LFDA,zheng2017unlabeled,lin2017improving} have been proposed. Mainstream works can be roughly categorized into two groups as follows.

The first group of methods focus on designing discriminative and invariant features \cite{LOMOXQDA,SalientColorName,GaussianReIDDescriptor,Invariant_Color_FeaturesReID,VIPeR,SalienceLearning,WhosDescriptors}. Earlier works include fisher vector based local descriptors \cite{ReIDFisherVector}, color invariant features \cite{Color_Invariants_ReID} and saliency learning based methods \cite{SalienceLearning}. Recently, some new proposed descriptors have gained good performance, i.e., salient color names \cite{SalientColorName}, local maximal occurrence (LOMO) feature \cite{LOMOXQDA}, weighted histograms of overlapping Stripes (Whos) \cite{WhosDescriptors}, and Gaussian of Gaussian (GOG) descriptor \cite{GaussianReIDDescriptor}. The GOG descriptor is used in this work. It describes a local region in a person image via hierarchical Gaussian distribution in which both means and covariances are included in their parameters. Specifically, it models the region as a set of multiple Gaussian distributions in which each Gaussian represents the appearance of a local patch. The characteristics of the set of Gaussian distributions are again described by another Gaussian distribution. The final descriptor fuses multiple GOG descriptor extracted from different color spaces. It performs well on a lot of re-ID datasets. 

The second group of methods aim to learn a robust and discriminative distance function for recognizing people across views \cite{ChenDaPeng_similarityLearning,KISSme,KLFDA,LADFReID,LFDA,LOMOXQDA,MLAPG,MetricEnsemble,MultiTaskDMLReID,PCCA,zhang2016learning}.
 In this group, some works aim to learn a Mahalanobis-like distance metric \cite{MLAPG,PRID_MDLReID,ERMML+}, while some methods focus on seeking a discriminative projection. \cite{zhang2016learning,KLFDA,RCCA,ZhengWeishi_AsymmetricDistance}.
These two subgroups actually are closely related.
We briefly introduce some well-known works as follows. 
Liao et al. \cite{MLAPG} proposed a logistic metric learning approach with PSD constraints and asymmetric sample weight strategy. 
Zheng et al. \cite{WeishiZheng_RelativeDistanceReID} formulated re-ID as a relative distance comparison learning problem by maximizing the probability that relevant samples have smaller distance than the irrelevant ones. 
Kostinger et al. \cite{KISSme} designed a simple and effective metric learning method by computing the difference between the intra-class and inter-class covariance matrix, while the presented algorithm is very sensitive to the dimension of feature representation.  
As an improvement, Liao et al. \cite{LOMOXQDA} proposed a cross-view quadratic discriminant analysis (XQDA) method by learning a more discriminative distance metric and a low-dimensional subspace simultaneously. 
Pedagadi et al. \cite{LFDA} applied the local fisher discriminant analysis algorithm to match person images by maximizing the inter-class separability while preserving the multi-class modality, whose kernel version was presented for re-ID in \cite{KLFDA}. 
Zhang et al. \cite{zhang2016learning} proposed to overcome the small-sample-size problem in re-ID by learning a discriminative null space, where the within-class scatter is minimized to zero while maximizing the relative between-class separation simultaneously.
Although lots of works have been developed, they are mainly developed in the fully-supervised setting. Once only a small number of labeled data are available, supervised methods are vulnerable to over-fitting. 
Therefore, the purpose of this work is to present a semi-supervised re-ID approach which can utilize the abundant unlabeled data to enhance the learning performance.

There are few semi-supervised re-ID methods, except for \cite{liu2014semi,IterativeLaplacian,SSMFL2013,karaman2014leveraging}.
Figueira et al. \cite{SSMFL2013} designed a semi-supervised method which exploits the general framework of multi-view learning with manifold regularization. It treats each person as a single class and poses re-ID as a multiple class recognition problem. However, conventional multiple class recognition methods may not be suitable for re-ID since each person usually has very limited images in re-ID.
Liu et al. \cite{liu2014semi}, proposed a semi-supervised coupled dictionary learning for re-ID, in which unlabeled data are used to improve the re-constructive ability of dictionaries. The authors assumed that the sparse representation coefficients of two matched images should be strictly equivalent, which is too strong to cope with dramatic changes of person's appearance.
Kodirov et al. \cite{IterativeLaplacian} presented a semi-supervised re-ID approach with graph Laplacian regularization, in which the visual similarity between a pair of unlabeled person images is computed using original low-level feature representation, which may result in a suboptimal performance. Different with \cite{IterativeLaplacian}, the proposed approach computes the similarity in a discriminative subspace learned using the available labeled data. Kodirov et al. \cite{kodirov2016person} also designed an unsupervised re-ID approach by introducing a new $\ell_1$-norm based graph Laplacian term instead of the conventional squared $\ell_2$-norm in \cite{IterativeLaplacian}, which can also be extended to a semi-supervised case. Karaman et al. \cite{karaman2014leveraging} described a semi-supervised re-ID approach. It combines discriminative models of person identity with a conditional random field to exploit the local manifold approximation induced by KNN graph. Different with our proposed approach, it is mainly designed for multi-shot scenarios where meaningful structure can be discovered easily. While, our approach can perform well in the challenging single-shot scenarios.
Zhang et al. \cite{zhang2016learning} introduced a semi-supervised extension for re-ID based on the self-training strategy \cite{zhu2009introduction}. It combines the pseudo-classes with the labeled data together into a new training set to learn the projection. Its difference with our work is that we place the pseudo pairwise information in a separate regularization term, which can reduce the negative effects of the incorrect matching pairs in the pseudo pairwise relationships and obtain more stable performance.

We also introduce a multi-kernel based extension in this work which exploits the complementation of multiple kernel representations. 
Some existing works \cite{MetricEnsemble,zhao2014learning} have investigated the effects of multiple feature representations for re-ID by a score-level fusion. Different with them, we focus on a kernel-level (feature-level) fusion.

Loy et al. \cite{loy2013person} formulated re-ID as a manifold ranking problem in an unsupervised way. It exploits the manifold structure revealed by a large quantity of gallery samples to obtain more robust ranking result. When combined with a distance metric learning method, it functions as a post-ranking approach.
In this work, we have investigated the effect of this manifold ranking approach. Experimental results demonstrate that our approach and the manifold ranking method can complement each other very well.

\subsection{Semi-supervised Learning}
Semi-supervised learning is a classic topic in the area of machine learning, which has numerous literatures. In this subsection, we only review some semi-supervised learning works which are based on self-training \cite{zhu2009introduction} or related to our work. 
 Self-training \cite{zhu2009introduction} is a commonly used semi-supervised learning technique and probably the earliest idea about using unlabeled data \cite{chapelle2009semi}. It is also known as self-learning, self-labeling, or decision-directed learning. This is a wrapper-algorithm that repeatedly uses a supervised learning method. It starts by training on the labeled data only. In each step a part of the unlabeled points are labeled according to the current decision function; then the supervised method is retrained using its own predictions as additional labeled points. 

\cite{mclachlan1975iterative} is one of the earliest works that applies this strategy to design an iterative reclassification procedure. \cite{yarowsky1995unsupervised} is a well-known example of self-training for word sense disambiguation. In \cite{basu2002semi}, this strategy is used for semi-supervised clustering. Rosenberg et al. \cite{rosenberg2005semi} applied self-training to object detection systems from images. Recently, self-training is explored in a state-of-the-art work \cite{loog2016contrastive} that proposes a general way to perform semi-supervised parameter estimation for likelihood-based classifiers and their estimates are never worse than the supervised solution in terms of the log-likelihood on the full training set.
 As a classic technique, self-training has been widely used for classification, clustering, regression, and other specific tasks in the past decades, and it still motivates researchers to develop new algorithms for specific applications nowadays. 

In this paper, we focus on designing a semi-supervised learning approach for person re-ID using self-training. Note that re-ID is a retrieval problem, and there is no intersection between the persons (classes) in training and testing. Therefore, most self-training based semi-supervised methods that are mainly designed for classification problems, are not suitable for re-ID. Owing to its simplicity and effectiveness for re-ID, subspace learning is explored as the basic learning method in our approach. Thereby, our approach is also related to
some semi-supervised subspace/metric learning approaches \cite{Cai_SSDiscriAnalysis,SSDMLImageRetrievalCVPR08,SSLFDA}. Our approach and  \cite{Cai_SSDiscriAnalysis,SSDMLImageRetrievalCVPR08} follow the same way to leverage unlabeled data, which encodes the neighborhood structure of unlabeled data in a Laplacian regularizer. Different with our approach,  \cite{Cai_SSDiscriAnalysis} is extended from linear discriminant analysis (LDA), while our approach seems more like a semi-supervised locality preserving projections (LPP) \cite{he2003locality}. \cite{SSDMLImageRetrievalCVPR08} focuses on learning a PSD constrained Mahalanobis metric and it uses a general loss term of metric learning. \cite{SSLFDA} is the semi-supervised extension of local fisher discriminant analysis (LFDA) \cite{sugiyama2006local} which combines the supervised LFDA and the unsupervised PCA to jointly exploit labeled and unlabeled data. It doesn't exploit the neighborhood structure of unlabeled data. Besides, to better leverage unlabeled data for re-ID, our approach uses self-training to repeatedly update the neighborhood structure of unlabeled data. Experimental results in this work have demonstrated that the self-training strategy significantly enhances the learning performance. 
Our main contribution is that we introduce a simple and effective re-ID approach which can exploit both labeled and unlabeled data using the self-training strategy.

\section{The Proposed Approach}
In this section, we describe our re-ID method. First, we briefly show how to perform re-ID in a fully-supervised subspace in Subsection \ref{Section3.1}. Then, we introduce the proposed semi-supervised case in detail in Subsection \ref{Section3.2}, followed by a multi-kernel based extension in Subsection \ref{Section3.3}.
\label{Section3}
\subsection{Person Re-ID in a Fully-supervised Subspace}\label{Section3.1}
In this work,  we formulate person re-ID as a subspace learning problem, which is similar with \cite{LFDA,KLFDA,RCCA}. 
Assume that we are given a set of \(n\) labeled training person images \(\mathbf{X}_l=\{{\mathbf{x}_i\}}_{i=1}^n\in \mathbb{R}^{d \times n}\), and their label set \(\mathbf{Y}_l=\{y_i\}_{i=1}^n\in \mathbb{R}^{n}\),
where $d$ denotes the dimension of feature vector. The task is to learn a squared distance function $d^2_\mathbf{U}(\mathbf{x}_i,\mathbf{x}_j)$ which is parameterized by a low-dimensional projection $\mathbf{U}$ defined as follow:
\begin{equation}\label{Eq1}
d^2_\mathbf{U}(\mathbf{x}_i,\mathbf{x}_j)=\left\Vert\mathbf{U}^\mathrm{T}\mathbf{x}_i-\mathbf{U}^\mathrm{T}\mathbf{x}_j\right\Vert^2,
\end{equation}
where $\mathbf{U}\in \mathbb{R}^{d \times r}(r\ll d)$ is a low-dimensional projection matrix which maps the person images from disjoint camera views into a common subspace where person re-ID can be performed easily. $r$ is the dimension of the projected subspace. The learned distance is expected to be small if $\mathbf{x}_i$ and $\mathbf{x}_j$ belong to the same person ($y_i=y_j$).
Under this expectation, we formulate re-ID as
\begin{equation}\label{Eq2}
\begin{aligned}
\mathbf{U}^*&=\arg \min\limits_{\mathbf{U}} \mathcal{L}\left(\mathbf{X}_l,\mathbf{U}, \mathbf{W}_l\right)\\
&=\frac{1}{2}\sum\limits_{i=1}^n \sum\limits_{j=1}^n  W^l_{ij}\left\Vert\mathbf{U}^\mathrm{T}\mathbf{x}_i-\mathbf{U}^\mathrm{T}\mathbf{x}_j\right\Vert^2,
\end{aligned}
\end{equation}
where $W^l_{ij}$ is an element of a weight matrix $\mathbf{W}^l\in \mathbb{R}^{n \times n}$ which encodes the pairwise constraints information between each pair of person images
\begin{equation}\label{Eq3}
W^l_{ij}=\left \{
\begin{array}{cc}
1& \textrm{if } y_i=y_j\\
0& \textrm{otherwise}
\end{array}
\right..
\end{equation}

The loss function $\mathcal{L}\left(\mathbf{X}_l,\mathbf{U},\mathbf{W}^l\right)$ can be rewritten as $\mathbf{tr}\left(\mathbf{U}^\mathrm{T}\mathbf{X}_l\mathbf{L}^l\mathbf{X}_l^\mathrm{T}\mathbf{U}\right)$,
where $\mathbf{tr(\cdot{•})}$ denotes the trace operator and  $\mathbf{L}^l=\mathbf{D}^l-\mathbf{W}^l$ is known as the graph Laplacian matrix. $\mathbf{D}^l$ is a diagonal matrix whose diagonal elements equal to the sums of the row entries of $\mathbf{W}^l$, i.e., $D^l_{ii}=\sum_j W^l_{ij}$.
By adding a constraint $\mathbf{tr}\left(\mathbf{U}^\mathrm{T}\mathbf{X}_l\mathbf{D}^l\mathbf{X}_l^\mathrm{T}\mathbf{U}\right)=1$, the minimization problem in Eq. (\ref{Eq2}) can be easily solved with a generalized eigen-decomposition. 
The final projection $\mathbf{U}^*=\left[\mathbf{u}_1,\mathbf{u}_2,\cdots,\mathbf{u}_r\right]\in \mathbb{R}^{d \times r}$ is constituted by the resulting eigenvectors associated to the $r$ smallest eigenvalues. Usually, $r$ is set to the difference of the number of labeled persons and one.

\subsection{Person Re-ID in a Self-trained Subspace}\label{Section3.2}
The above method seeks the discriminative projection based on merely labeled data. However, the main bottleneck for fully-supervised person re-ID is the limited availability of labeled training samples. When only a small number of labeled image pairs are available, the solution of above method tends to be over-fitted to the labeled data. In this subsection, we introduce a self-trained subspace learning framework for person re-ID in a semi-supervised setting.

Given a set of $n$ labeled person images \(\mathbf{X}_l=\{{\mathbf{x}_i\}}_{i=1}^n\in \mathbb{R}^{d \times n}\) and $u$ unlabeled person images \(\mathbf{X}_u=\{{\mathbf{x}_i\}}_{i=n+1}^{n+u}\in \mathbb{R}^{d \times u}\),  the task is to learn a projection $\mathbf{U}\in \mathbb{R}^{d \times r'}$ which has good generalization capability and discriminative power. A common way is to formulate the semi-supervised learning problem as the following general form:
\begin{equation}\label{Eq4}
\mathbf{U}^*=\arg \min\limits_\mathbf{U} \mathcal{L} \left(\mathbf{X}_l,\mathbf{U}, \mathbf{W}_l\right)+\eta \mathcal{R}\left(\mathbf{X}_u,\mathbf{U}\right),
\end{equation}
where the first term \(\mathcal{L} \left(\mathbf{X}_l,\mathbf{U},\mathbf{W}_l\right)\) is the labeled term in Eq. (\ref{Eq2}) which only relies on labeled data, and the second term $\mathcal{R}\left(\mathbf{X}_u,\mathbf{U}\right)$ is a regularization term constructed by unlabeled data. The trade-off between these two terms is captured by a small regularization parameter ($\eta>0$). The main problem is how to utilize the unlabeled data to construct the regularization term. A widely-used strategy in computer vision problems \cite{SSDMLImageRetrievalCVPR08,SSTransferMetric,IterativeLaplacian,JunYu_multiviewDML,Cai_SSDiscriAnalysis} is encoding the intrinsic geometric structure of unlabeled data into a regularizer under the manifold assumption that visually similar samples are more likely to share the same class label. A KNN graph is usually constructed to model the relationship between nearby data nodes, where an edge will be placed between two nodes if they are close.
\begin{algorithm}[t] %\small
\caption{The proposed self-trained subspace learning approach}
\begin{algorithmic}[1]
{
\INPUT {The labeled training data $\mathbf{X}_l$ and its weight matrix $\mathbf{W}^l$; the unlabeled training data $\mathbf{X}_u$; the parameter $\eta$; the maximal number of iterations $\mathcal{T}$.}
\Initialization  \(\mathbf{U}^0=\arg\min\limits_{\mathbf{U}} \mathcal{L}\left(\mathbf{X}_l,\mathbf{U},\mathbf{W}^l\right)\); $t=1$.\\
\STEPS $t=1, 2,\cdots, \mathcal{T}$\\
\STATE Project $\mathbf{X}_u$ into a low-dimensional subspace through $\mathbf{Z}^t_u=(\mathbf{U}^{t-1})^\mathrm{T}\mathbf{X}_u$;\\
\STATE Build the pseudo pairwise relationships among the unlabeled data by constructing a KNN graph using $\mathbf{Z}^t_u$;\\
\STATE Encode the pseudo pairwise relationships into a weight matrix $\mathbf{W}^{u,t}$;\\
\STATE Solve $\mathbf{U}^t\!=\!\arg\min\limits_\mathbf{U} \mathcal{L}\left(\mathbf{X}_l,\!\mathbf{U},\!\mathbf{W}^l\right)\!+\!\eta \mathcal{R}\left(\mathbf{X}_u,\!\mathbf{U},\!\mathbf{W}^{u,t}\right)$ to obtain the new projection matrix;\\
\NewUNTIL {$t>\mathcal{T}$ or the stop condition is met}.
\OUTPUT {The projection matrix $\mathbf{U}$.}\\
}
\label{Algorithm1}
\end{algorithmic}
\end{algorithm}
Nearest neighbors selection is performed by computing the Euclidean distance between node $i$ and node $j$ using original feature representation.
However, the original feature representation has very low discriminative power due to the dramatic appearance change across camera-views in person re-ID. As a result, the constructed KNN graph may be misleading, which will degrade the learning performance.

To overcome the above problem, in this work, we introduce self-training to better utilize the unlabeled data $\mathbf{X}_u$. We first apply the fully-supervised method described in Subsection \ref{Section3.1} to learn an initial projection matrix $\mathbf{U}^0$ using the labeled data $\mathbf{X}_l$ only. Then, we project the unlabeled data $\mathbf{X}_u$ into a low-dimensional subspace using $\mathbf{U}^0$.
The low-dimensional representations $\mathbf{Z}_u=(\mathbf{U}^0)^\mathrm{T}\mathbf{X}_u$ are used to obtain the cross-view adjacency relationships among the unlabeled samples $\mathbf{X}_u$ by constructing a KNN graph. The cross-view adjacency relationships can be viewed as a kind of pseudo pairwise relationships. Given the pseudo pairwise relationships, we can leverage the unlabeled data in a supervised way. We encode the pseudo pairwise relationships into the following weighted matrix $\mathbf{W}^u\in \mathbb{R}^{u \times u}$ for $\mathbf{X}_u$:
\begin{equation}\label{Eq5}
W^u_{ij}=\left \{
\begin{array}{cc}
1 & \textrm{if }\mathbf{x}_{i} \in {\mathcal{N}\left(\mathbf{x}_{j}\right)} \;\, \textrm{or} \quad\mathbf{x}_{j} \in {\mathcal{N}\left(\mathbf{x}_{i}\right)}\\
0 & \textrm{otherwise}
\end{array}
\right.,
\end{equation}
where $\mathbf{x}_{i}$ and $\mathbf{x}_{j}$ are two unlabeled person images from different views, $n+1\leq i, j\leq {n+u}$. $\mathcal{N}\left(\mathbf{x}_{i}\right)$ denotes the nearest neighbor list of $\mathbf{x}_{i}$.
After obtaining the weighted matrix $\mathbf{W}^u$, the regularization term in Eq. (\ref{Eq4}) is constructed as
\begin{equation}\label{Eq6}
\begin{aligned}
\mathcal{R}\left(\mathbf{X}_u, \! \mathbf{U},\! \mathbf{W}^u\right)
&=\frac{1}{2}\!\!\sum\limits_{i=n+1}^{n+u}\sum\limits_{j=n+1}^{n+u} \!\!\!{W^u_{ij}\left\Vert\mathbf{U}^\mathrm{T}\mathbf{x}_i \!-\! \mathbf{U}^\mathrm{T}\mathbf{x}_j\right\Vert^2}\\
&=\mathbf{tr}\left(\mathbf{U}^\mathrm{T}\mathbf{X}_u\mathbf{L}^u\mathbf{X}_u^\mathrm{T}\mathbf{U}\right)
\end{aligned}
\end{equation}
where $\mathbf{L}^u=\mathbf{D}^u-\mathbf{W}^u$ is the Laplacian matrix for unlabeled data. $\mathbf{D}^u$ is a diagonal matrix whose diagonal elements equal to the sum of the rows entries of $\mathbf{W}^u$. Hence, by utilizing the pseudo pairwise relationships for unlabeled data, we rewrite the semi-supervised learning problem in Eq. (\ref{Eq5}) as
\begin{equation}\label{Eq7}
\begin{aligned}
\mathbf{U}^*=&\arg\min\limits_\mathbf{U}{\mathcal{L}\left(\mathbf{X}_l,\!\mathbf{U},\!\mathbf{W}^l\right)}\!+\!\eta\mathcal{R}\left(\mathbf{X}_u,\!\mathbf{U},\!\mathbf{W}^u\right)\\
=&\arg\min\limits_\mathbf{U}\mathbf{tr}{\Big(\mathbf{U}^\mathrm{T}\!\!\left(\mathbf{X}_l\mathbf{L}^l\mathbf{X}_l^\mathrm{T}\!+\!\eta\mathbf{X}_u\mathbf{L}^u\mathbf{X}_u^\mathrm{T}\right)\!\mathbf{U}\Big)}\\
s.t. &\quad \mathbf{tr}{\Big(\mathbf{U}^\mathrm{T}\!\!\left(\mathbf{X}_l\mathbf{D}^l\mathbf{X}_l^\mathrm{T}\!+\!\eta\mathbf{X}_u\mathbf{D}^u\mathbf{X}_u^\mathrm{T}\right)\!\mathbf{U}\Big)}=1
\end{aligned}.
\end{equation}
We can easily solve the minimization problem in Eq.(\ref{Eq7}) with a generalized eigen-decomposition to obtain the projection matrix. Note that, to obtain more stable performance, the dimension $r'$ of the final subspace is fixed as the difference of the number of labeled persons and one. In other words, we do not increase the dimension of the subspace although we have exploited abundant unlabeled data.

It can be seen from the above analysis that the pseudo pairwise relationships among the unlabeled persons play a crucial role in the proposed approach. However, it inevitably includes some mismatching pairwise relationships due to the complex viewpoint variations. Therefore, given the newly learned projection, we refine the pseudo pairwise relationships and relearned the discriminative projection with these updated pseudo pairwise relationships. This procedure is iterated until the pseudo pairwise relationships remain unchanged.
We summarize the proposed self-trained subspace learning algorithm in Algorithm \ref{Algorithm1}.

\subsection{Multi-kernel based Extension} \label{Section3.3}
To better exploit the non-linearity in person's appearance and the complementary information shared by multiple kernels, we employ the multi-kernel embedding \cite{shrivastava2014multiple,Shao2016} for the proposed approach. The task is to learn a kernelized projection $\mathbf{P}$. Given $M$ kernel matrices with the same size $\mathcal{K}_1,\cdots,\mathcal{K}_M\in\mathbb{R}^{(n+u)\times (n+u)}$ constructed using the total training set $\mathbf{X}=\{\mathbf{x}_i\}_{i=1}^{n+u}$, the primary task is to learn a fused kernel matrix
\begin{equation}\label{Eq8}
\mathcal{K}=\sum\limits_{m=1}^{M} \beta_m \mathcal{K}_m, s.t. \sum\limits_{m=1}^{M} \beta_m =1, \beta_m>0,
\end{equation}
where $\beta_m$ is a non-negative weight for the kernel matrix $\mathcal{K}_m$. In this formulation, $\{\mathcal{K}_m\}_{m=1}^M$ can be the same classic kernels with different hyper-parameters, different feature descriptors or different kernels. Let $\phi_m(\cdot)$ be the feature map for $\mathcal{K}_m$. Then $\mathcal{K}_m$ can be expressed by an inner product in the kernel space $\mathcal{K}_m\!\!=\!\!\phi_m(\mathbf{X})^\mathrm{T}\phi_m(\mathbf{X})$, where  $\phi_m(\mathbf{X})=[\phi_m(\mathbf{x}_1),\cdots,\phi_m(\mathbf{x}_n),\cdots,\phi_m(\mathbf{x}_{n+u})]$. We can rewrite the fused kernel matrix in Eq. (\ref{Eq8}) as
\begin{equation}\label{Eq9}
\mathcal{K}=\sum\limits_{m=1}^{M}{\beta_m\phi_m(\mathbf{X})^\mathrm{T}\phi_m(\mathbf{X})}=\phi(\mathbf{X})^\mathrm{T}\phi(\mathbf{X}),
\end{equation}
where $\phi(\cdot)=\left[\sqrt{\beta_1}\phi_1(\cdot)^\mathrm{T},\cdots,\sqrt{\beta_M}\phi_M(\cdot)^\mathrm{T}\right]^\mathrm{T}$ is the fused feature map.
We employ the kernel alignment \cite{qiu2009framework} approach to decide the kernel weights.
\begin{equation}\label{Eq10}
\beta_m=\frac{A(\mathcal{K}_m^{ll},\mathcal{K}d)}{\sum_{{m'}=1}^{M}{A(\mathcal{K}_{m'}^{ll},\mathcal{K}d)}}.
\end{equation}
where $\mathcal{K}d\in\mathbb{R}^{n\times n}$ is the ideal kernel matrix for the labeled person samples, whose element is 1 where rows and columns correspond to the same person or 0 everywhere else. $\mathcal{K}_{m'}^{ll}\in\mathbb{R}^{n\times n}$ is a part of the base kernel matrix $\mathcal{K}_{m'}$, which corresponds to the labeled person samples. The alignment score between two kernel matrices is defined as follows
\begin{equation}\label{Eq11}
A(\mathcal{K}_{m'}^{ll},\mathcal{K}d)=\frac{\langle \mathcal{K}_{m'}^{ll},\mathcal{K}d \rangle_F}{\sqrt{{\langle \mathcal{K}_{m'}^{ll},\mathcal{K}_{m'}^{ll} \rangle_F}{\langle \mathcal{K}d,\mathcal{K}d \rangle_F}}},
\end{equation}
where ${\langle \mathcal{K}_{m'}^{ll},\mathcal{K}d \rangle_F}=\textbf{tr}((\mathcal{K}_{m'}^{ll})^\mathrm{T}\mathcal{K}d)$.

\begin{figure*}[tbp]
	\begin{center}
		\includegraphics[width=5.8 in]{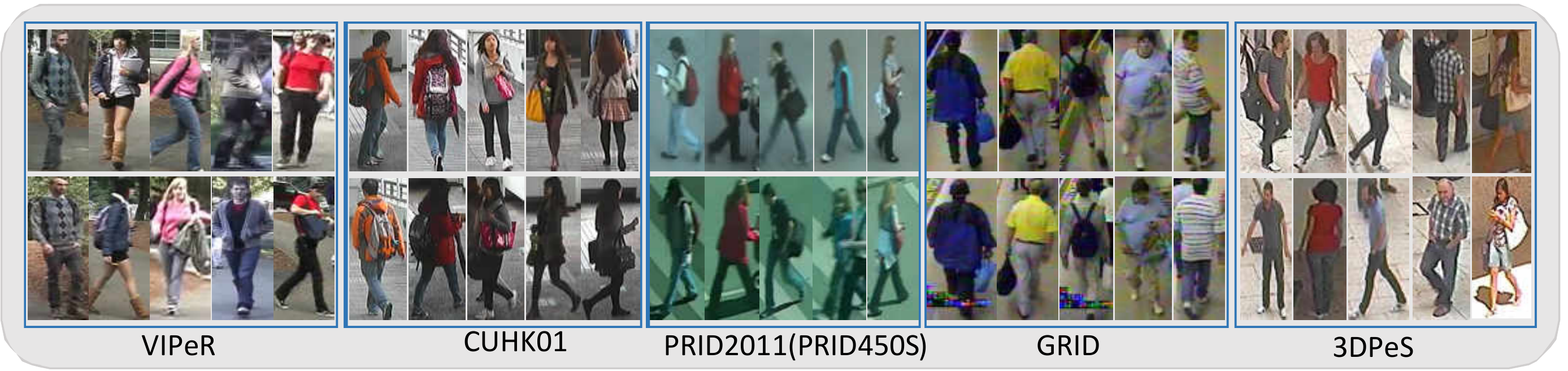}
	\end{center}
	\vspace{-0.15in}
	\caption{Sample images from six person re-identification datasets. From left to right: VIPeR, CUHK01, PRID2011 (PRID450S), GRID, and 3DPeS.}
	\vspace{-0.15in}
	\label{figure2}
\end{figure*}

When $\mathcal{K}$ and $(\beta_1,\beta_2,\cdots,\beta_M)$ are obtained, we denote
$\mathcal{K}=\Big[
\begin{array}{ll}
\mathcal{K}^{ll}  &\mathcal{K}^{lu} \\
\mathcal{K}^{ul} &\mathcal{K}^{uu}
\end{array}
\Big]$, $\mathcal{K}^{l}=\Big[
\begin{array}{c}
\mathcal{K}^{ll}\\
\mathcal{K}^{ul}
\end{array}
\Big]$, and $\mathcal{K}^{u}=\Big[
\begin{array}{c}
\mathcal{K}^{lu}\\
\mathcal{K}^{uu}
\end{array}
\Big]$. We denote the fused feature maps of the labeled training set and the unlabeled training set as
 $\phi(\mathbf{X}_l)=[\phi(\mathbf{x}_1),\cdots,\phi(\mathbf{x}_n)]$, and $\phi(\mathbf{X}_u)=[\phi(\mathbf{x}_{n+1}),\cdots,\phi(\mathbf{x}_{n+u})]$, respectively. The minimization problem in Eq. (\ref{Eq7}) can be solved by computing the following eigenvalue problem in the fused kernel space
\begin{equation}\label{Eq12}
\begin{aligned}
&\left(\phi(\mathbf{X}_l)\mathbf{L}^l\phi(\mathbf{X}_l)^\mathrm{T}+\eta \phi(\mathbf{X}_u)\mathbf{L}^u\phi(\mathbf{X}_u)^\mathrm{T}\right)\mathbf{u}\\
=&\lambda
\left(\phi(\mathbf{X}_l)\mathbf{D}^l\phi(\mathbf{X}_l)^\mathrm{T}+\eta \phi(\mathbf{X}_u)\mathbf{D}^u\phi(\mathbf{X}_u)^\mathrm{T}\right)\mathbf{u}
\end{aligned}.
\end{equation}
Since the eigenvectors are the linear combinations of $\phi(\mathbf{x}_1)$, $\phi(\mathbf{x}_2)$, $\cdots$, $\phi(\mathbf{x}_{n+u})$, there exists coefficients $p_i$ such that $\mathbf{u}=\sum_{i=1}^{n+u}{p_i \phi(\mathbf{x}_i)}=\phi(\mathbf{X})\mathbf{p},$
where $\mathbf{p}=\left[p_1,p_2,\cdots,p_{n+u}\right]^\mathrm{T}\in \mathbb{R}^{n+u}$. 

By simple algebra formulation, we can finally obtain the following kernelized eigenvalue problem
\begin{equation}\label{Eq13}
\begin{aligned}
&\left(\mathcal{K}^{l}\mathbf{L}^l(\mathcal{K}^{l})^\mathrm{T}+\eta
\mathcal{K}^{u}\mathbf{L}^u(\mathcal{K}^{u})^\mathrm{T} \right)\mathbf{p}\\
=&\lambda
\left(\mathcal{K}^{l}\mathbf{D}^l(\mathcal{K}^{l})^\mathrm{T}+\eta
\mathcal{K}^{u}\mathbf{D}^u(\mathcal{K}^{u})^\mathrm{T} \right) \mathbf{p}.
\end{aligned}
\end{equation}
Usually, it is common to apply a regularization technique for the eigenvalue problem to avoid the singularity of matrix. For simplicity, we denote \(\mathbf{A}=\mathcal{K}^{l}\mathbf{D}^l(\mathcal{K}^{l})^\mathrm{T}+\eta
\mathcal{K}^{u}\mathbf{D}^u(\mathcal{K}^{u})^\mathrm{T}\). We regularize $\bf{A}$ by adding an identity matrix, i.e., $\mathbf{A}=\mathbf{A}+\vartheta \frac{\mathbf{tr}(\mathbf{A})}{n+u}\mathbf{I}$, where $\mathbf{I}\in \mathbb{R}^{(n+u)\times (n+u)}$ is an identity matrix and $\vartheta$ is a small positive parameter. The final kernelized projection $\mathbf{P}=\left[\mathbf{p}_1,\cdots,\mathbf{p}_{r'}\right]\in \mathbb{R}^{(n+u)\times {r'}}$ is constituted by the resulting eigenvectors associated to the $r'$ smallest eigenvalues.

\section{Experimental Results and Analyses}
In this section, we first introduce the datasets, the evaluation protocol, and the experimental setting. Then, we evaluate the performance of our approach on multiple re-ID datasets.

\subsection{Datasets,  Evaluation protocol, and Setting}
\subsubsection{Datasets}
In this work we use six popular person re-ID datasets: VIPeR \cite{VIPeR}, CUHK01 \cite{li2012human},  and PRID2011 \cite{hirzer2011person}, PRID450S \cite{PRID_MDLReID}, GRID \cite{GRID}, and 3DPeS \cite{baltieri2011_308}.
Table \ref{Table1} provides a statistical summary of each dataset. In Table \ref{Table1}, we indicate the number of people, bounding boxes (BBoxes), distractors, and cameras (Cam) in each dataset. Fig. \ref{figure2} shows some sample images from these six datasets.

VIPeR \cite{VIPeR} is the most commonly-used dataset containing 632 persons in which each person has a pair of images taken from widely differing views. The large viewpoint change of 90 degrees or more as well as huge lighting variations make it one of the most challenging datasets. CUHK01 \cite{li2012human} is one of the largest benchmarks. It contains 971 persons from two disjoint camera views, where each person has two images in each camera view. It contains 3884 images in total. PRID2011 \cite{hirzer2011person} consists of person images recorded from two cameras (camera A and camera B) captures 385 persons and 749 persons, respectively, only 200 persons appear in both camera views. PRID450S \cite{PRID_MDLReID} is an extension of PRID2011. It contains 450 persons in which each person has a pair of images taken from two disjoint camera views. GRID \cite{GRID} has 250 image pairs collected from 8 non-overlapping cameras. 775 non-paired people are also included as distractors in the gallery set, which makes it extremely challenging. GRID suffers from viewpoint variations, background clutter, occlusions and low-resolution. 3DPeS is a set of selected snapshots of the original video dataset \cite{baltieri2011_308}, containing 192 people and 1011 images.

\subsubsection{Evaluation protocol}
For all datasets, all the individuals are randomly divided into two subsets, so that the training and testing sets contain half of the available individuals with no overlap on person identities. The single-shot experiment setting \cite{GaussianReIDDescriptor} is used for all datasets. We also report the multi-shot matching results for CUHK01. As random selection is involved, the evaluation procedure is repeated for 10 times and the mean results for all datasets are reported. We adopt the data splits in \cite{GaussianReIDDescriptor} for VIPeR, CUHK01, PRID450S, and GRID. We use the data splits in \cite{lisanti2014matching} for PRID2011, in which 100 persons are randomly selected for training from the 200 available persons present in both views in each data split and the remaining 100 persons of camera view A are used as probe set and the remaining 649 persons of another camera view are used as gallery set. We use the data splits in \cite{KLFDA} for 3DPeS, in which we randomly select one image of each individual in the testing set as gallery image, and the rest are used as probe images. The Cumulated Matching Characteristics (CMC) curve is used to evaluate the performance of all methods. It provides a ranking for every image in the gallery with respect to the probe.
\begin{table}[tbp]
	\caption{The characteristics of six person re-ID datasets. }{
		\renewcommand\arraystretch{1.2}
		\begin{tabular}{l|c|c|c|c}
			\hline
			\textbf{Datasets}      & \textbf{\# People}& \textbf{\# BBoxes} & \textbf{\# Distractors} & \textbf{\# Cam}  \\
			\hline
			VIPeR \cite{VIPeR}         &632 &1264 &0 &2 \\
			CUHK01\cite{li2012human}    &971 & 3884&0 & 2\\
			PRID2011 \cite{hirzer2011person}&200 &849 &649 &2 \\
			PRID450S\cite{PRID_MDLReID} &450 & 900& 0&2 \\
			GRID\cite{GRID}          &250  &500 &775 & 8\\
			3DPeS \cite{baltieri2011_308} &192 &1011 &0 & 8\\
			\hline		  			
	\end{tabular}}
	\label{Table1}
\end{table}	

\begin{figure*}[tbp]
	\begin{minipage}[b]{0.33\textwidth}
		\includegraphics[width=2.3in]{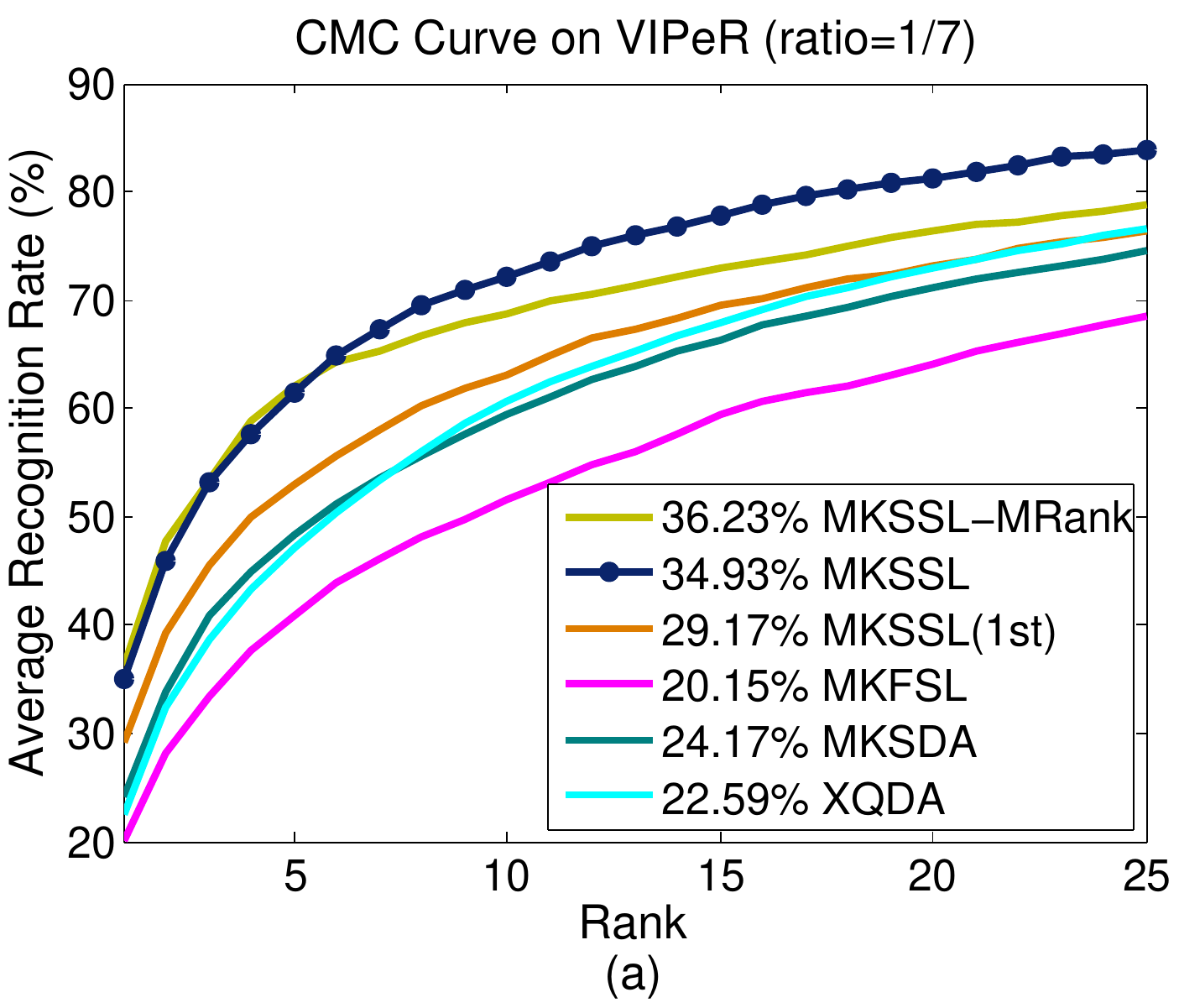} 
	\end{minipage}
	\begin{minipage}[b]{0.33\textwidth}
		\includegraphics[width=2.3in]{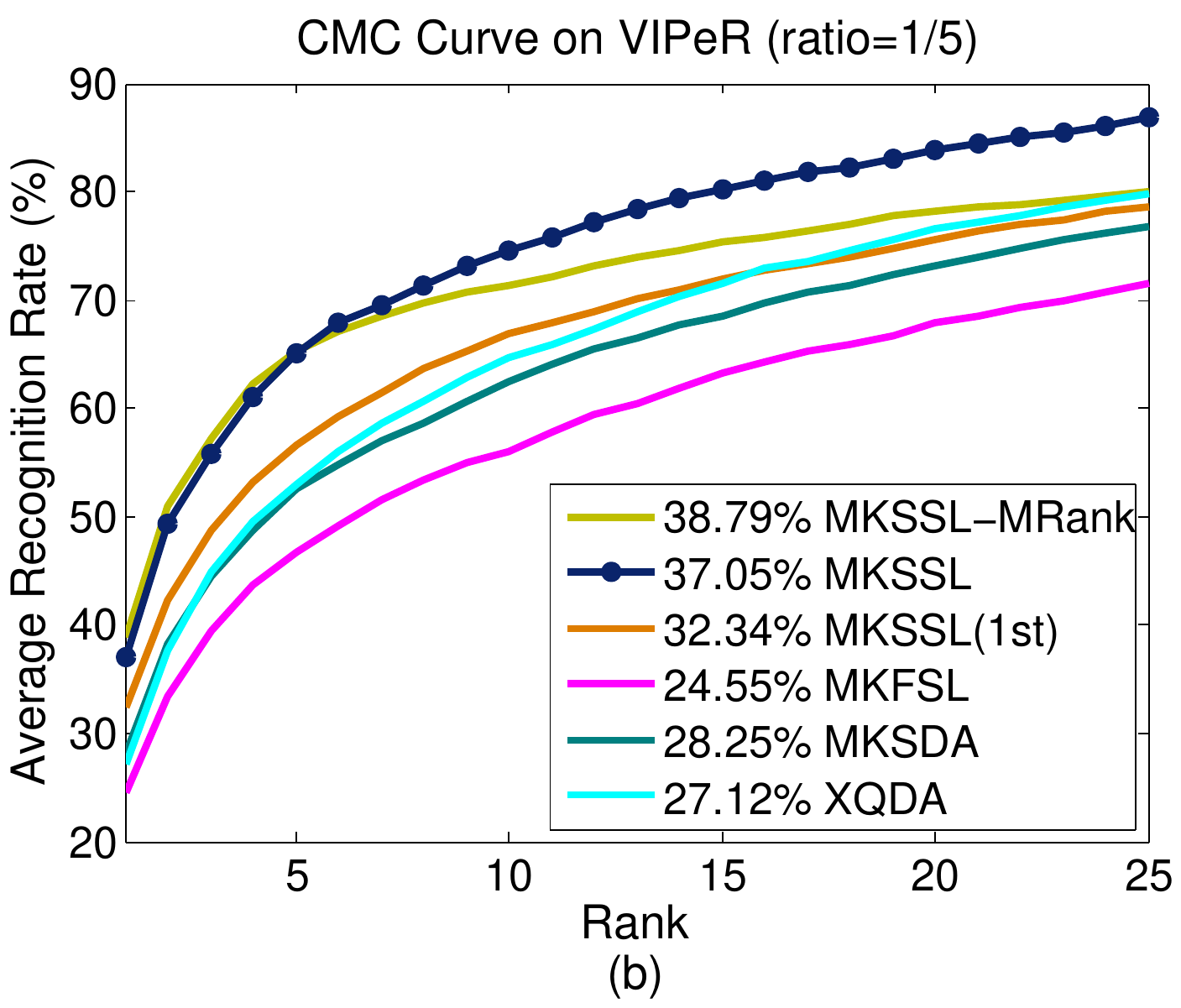} 
	\end{minipage}
	\begin{minipage}[b]{0.33\textwidth}
		\includegraphics[width=2.3in]{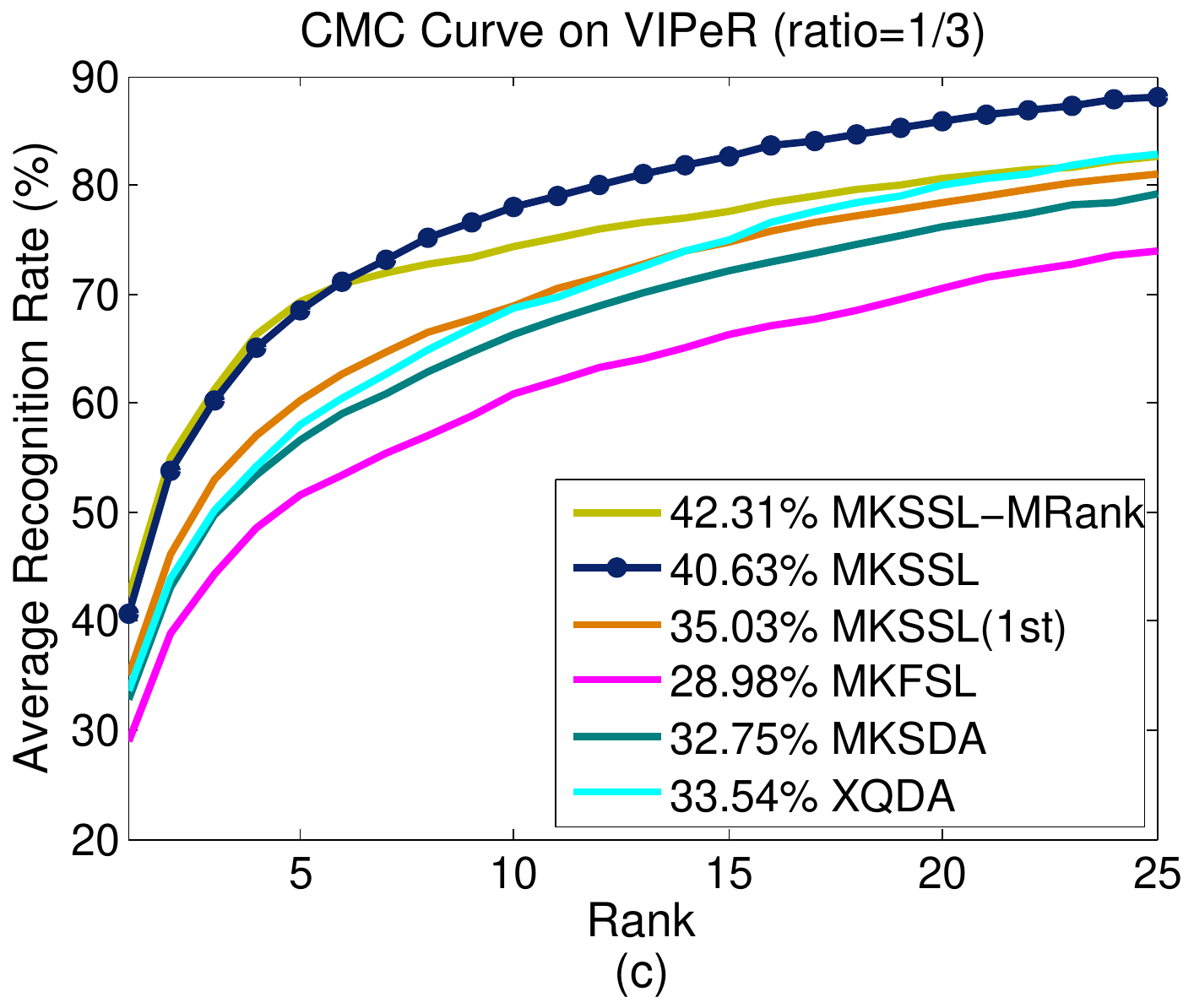} 
	\end{minipage}
	\vspace{-0.25in}
	\caption{Performance comparison of the proposed approach with different baselines on the VIPeR dataset with different settings of $ratio$. Rank-1 recognition rates are shown in the legends.  
	}
	\vspace{-0.20in}
	\label{figure3}
\end{figure*}
\subsubsection{Setting} \label{setting}
For the multi-kernel setting, we use the GOG descriptor \cite{GaussianReIDDescriptor} to construct 11 Gaussian kernels $\exp(-\frac{1}{c\mu}\Vert \mathbf{x}_i-\mathbf{x}_j \Vert^2)$, where $c$ varies from 2 to 3 with step 0.1 and $\mu$ is average squared Euclidean distance.

Parameter $\vartheta$ is set to 0.01. Two main parameters are $\eta$ in Eq. (\ref{Eq4}) and the number of nearest neighbors $k$ in KNN graph. We set them as $\eta=1$ and $k=2$ by cross-validation on the training set of CUHK01 and fix them for all datasets. The maximal iteration number $\mathcal{T}$ is set to 10. The iteration procedure will stop when the neighborhood structure of unlabeled data remains unchanged or changes very little. 

For the semi-supervised setting, we denote the proportion of labeled data in the training set as $ratio$ and the rest are used as unlabeled data. We evaluate the effectiveness of our approach with different settings of $ratio$. 
The dimension $r'$ ($r$) of the final subspace is fixed as the difference of the number of labeled persons and one.

The proposed approach is termed as MKSSL which includes a multi-kernel embedding. The baseline methods are constructed as follows:
\begin{itemize}
\item MKSSL(1st): The proposed approach with only one iteration.
\item MKFSL: The fully-supervised subspace learning method described in Subsection \ref{Section3.1} with the introduced multi-kernel embedding technique. Only the labeled data in the training set are used.
\item MKSDA: The classic semi-supervised discriminant analysis method in \cite{Cai_SSDiscriAnalysis} with the introduced multi-kernel embedding technique.
\item XQDA: The state-of-the-art fully-supervised metric learning method in \cite{LOMOXQDA}. Only the labeled data in the training set is used. 
\end{itemize}	

In addition, in the testing stage the most common way is directly performing image matching using Euclidean distance in the learned distance space, which is also the default strategy of MKSSL and other compared methods. In this work, we also integrate the manifold ranking method \cite{loy2013person} with the learned distance function of MKSSL in the testing stage to compute the similarity score between a probe image and a gallery image. We term this approach as MKSSL-MRank, in which the neighborhood graph of the manifold ranking method is constructed in the learned distance space of MKSSL instead of the original feature space. The performance of MKSSL-MRank will be evaluated in the following experiments. Note that, if not mentioned, our method and the above listed baseline methods all use the GOG descriptor in the following experiments.
 \begin{table}[htbp]\scriptsize
 						\caption{Performance comparison (CMC@rank-r, $\%$) of our approach with the reported results of state-of-the-art semi-supervised or fully-supervised methods on the VIPeR dataset.
 						 A larger number indicates a better result. }
 						\renewcommand\arraystretch{1.2}
 								\begin{center}
 			\begin{tabular}{c|l|c|c|c|c}
 										\hline 	
 		     \multicolumn{2}{c|} {\textbf{VIPeR}}     & r=1 & r=5 & r=10 & r=20 \\
 					    \hline
 				  \multirow{7}{*}{ {Semi-supervised}} & MKSSL-MRank [Ours]& 42.3  & 69.4 & 74.4  & 80.6 \\
 				   & MKSSL [Ours]& 40.6  & 68.5  & 78.1 & 85.9 \\
 				   & LOMO+MKSSL [Ours]& 31.2  & 52.5  & 62.9 & 72.8 \\
 								  \cline{2-6}
 				  & LOMO+LDNS \cite{zhang2016learning} & 31.7 & 59.4 & 72.8 & 84.9\\
 				  { $ratio=1/3$}   & SSMFL \cite{SSMFL2013}& 22.5 & 44.4 & 55.9 & 70.7  \\
 				   & DLIterLap \cite{IterativeLaplacian} & 32.5 & 61.8 & 74.3 & 84.1 \\
 				   & SSCDL \cite{liu2014semi} & 25.6 & 53.7 & 68.1 & 83.6 \\
 				  \hline\hline
 				  \multirow{12}{*}{} & MKFSL [Ours]& 51.0 & 81.4& 89.3 & 95.3 \\ 		  \cline{2-6}
 				  								& GOG+XQDA \cite{GaussianReIDDescriptor} & 49.7& 79.7 & 88.7 & 94.5 \\
 				  								& LOMO+LDNS  \cite{zhang2016learning}         & 42.3 & 71.5 & 82.9& 92.0 \\
 				  							   & LOMO+LSSCDL  \cite{HuchuanLu_Sample-Specific_SVM_Learning}& 42.7 &- & 84.3& 91.9\\
 				  					           & Ensemble \cite{MetricEnsemble}       & 45.9 & 77.5 & 88.9 & 95.8\\
 	 {Fully-supervised}&  LOMO+XQDA \cite{LOMOXQDA}                     & 40.0 &- & 80.5 & 91.1\\
 	 { ${ratio=1}$}	&  Semantic  \cite{shi2015transferring}             & 41.6 & 71.9 & 86.2 & 95.1\\
 				  					          &  LOMO+MLAPG \cite{MLAPG}		                            & 40.7 &- & 82.3 & 92.4\\	
 				  					          &  MTL-LORAE \cite{su2015multi}                    & 42.3 & 72.2 & 81.6 & 89.6\\
 				  					          & DLIterLap \cite{IterativeLaplacian}            & 38.9 & 70.8 & 78.5 & 86.1\\
 				  					          & UnL1Graph \cite{kodirov2016person}     & 41.5 & - & - & -\\
 				  					          & MCKCCA\cite{lisanti2016mckcca} & 47.9& -& 87.3 & 93.8\\
 					\hline			
 	\end{tabular}
 			\end{center}
 			\vspace{-0.20in}
 		\label{Table2}
 		\end{table}	
 	
 		\begin{figure*}[tbp] 
 		\begin{minipage}[b]{0.31\textwidth}
 			\includegraphics[width=2.2in]{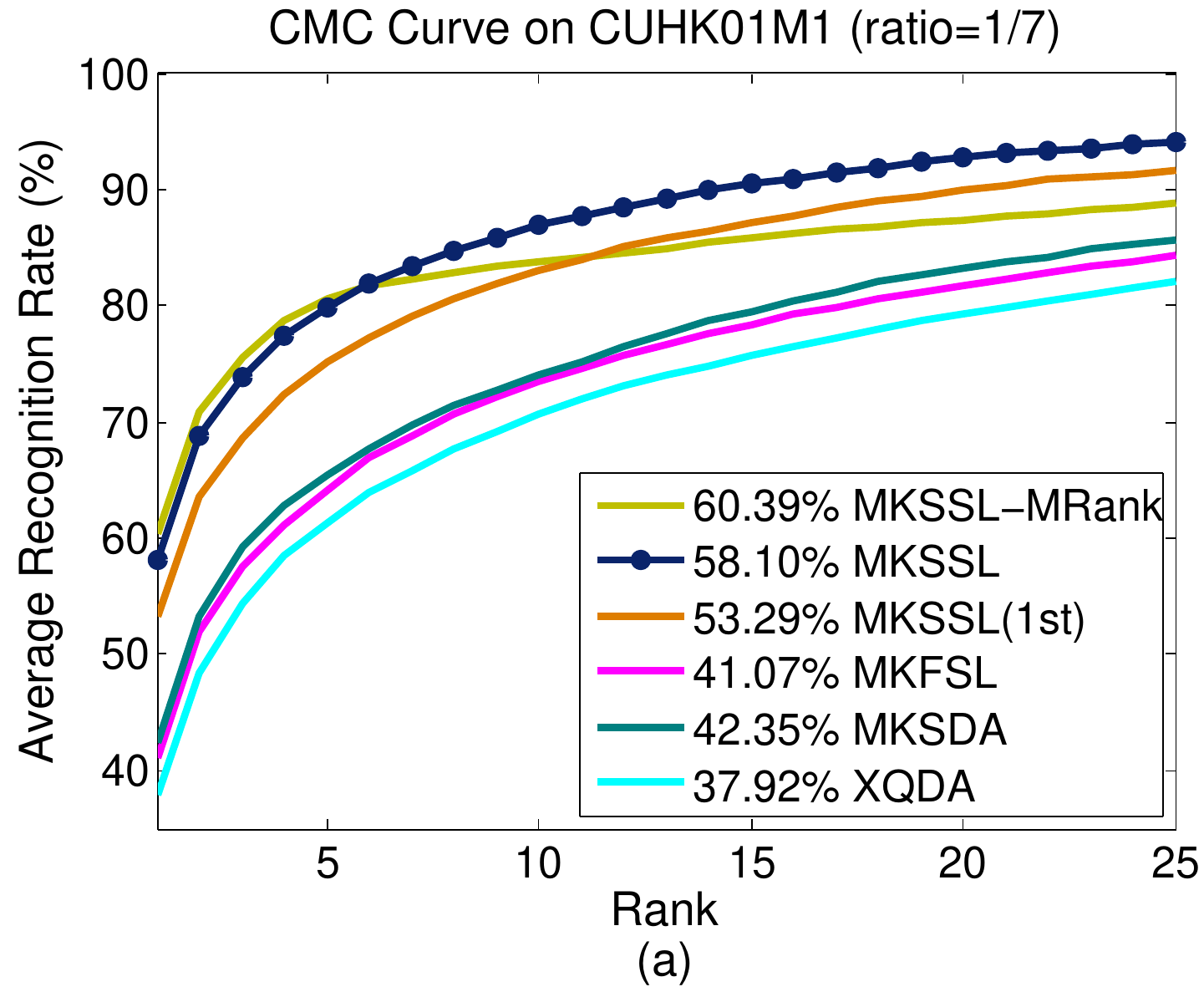} \\
 			\includegraphics[width=2.2in]{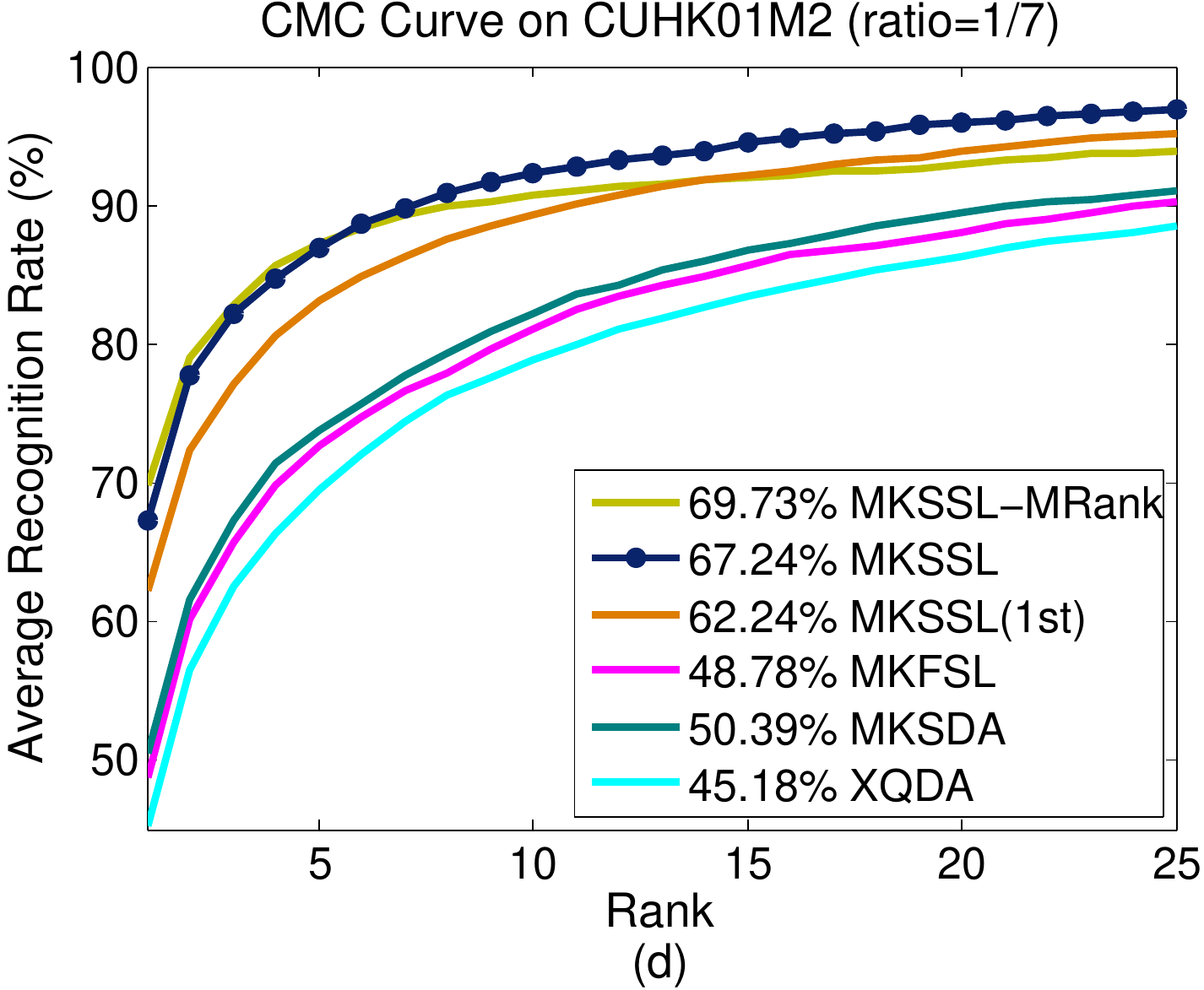} 
 		\end{minipage}
 		\begin{minipage}[b]{0.31\textwidth}
 			\includegraphics[width=2.2in]{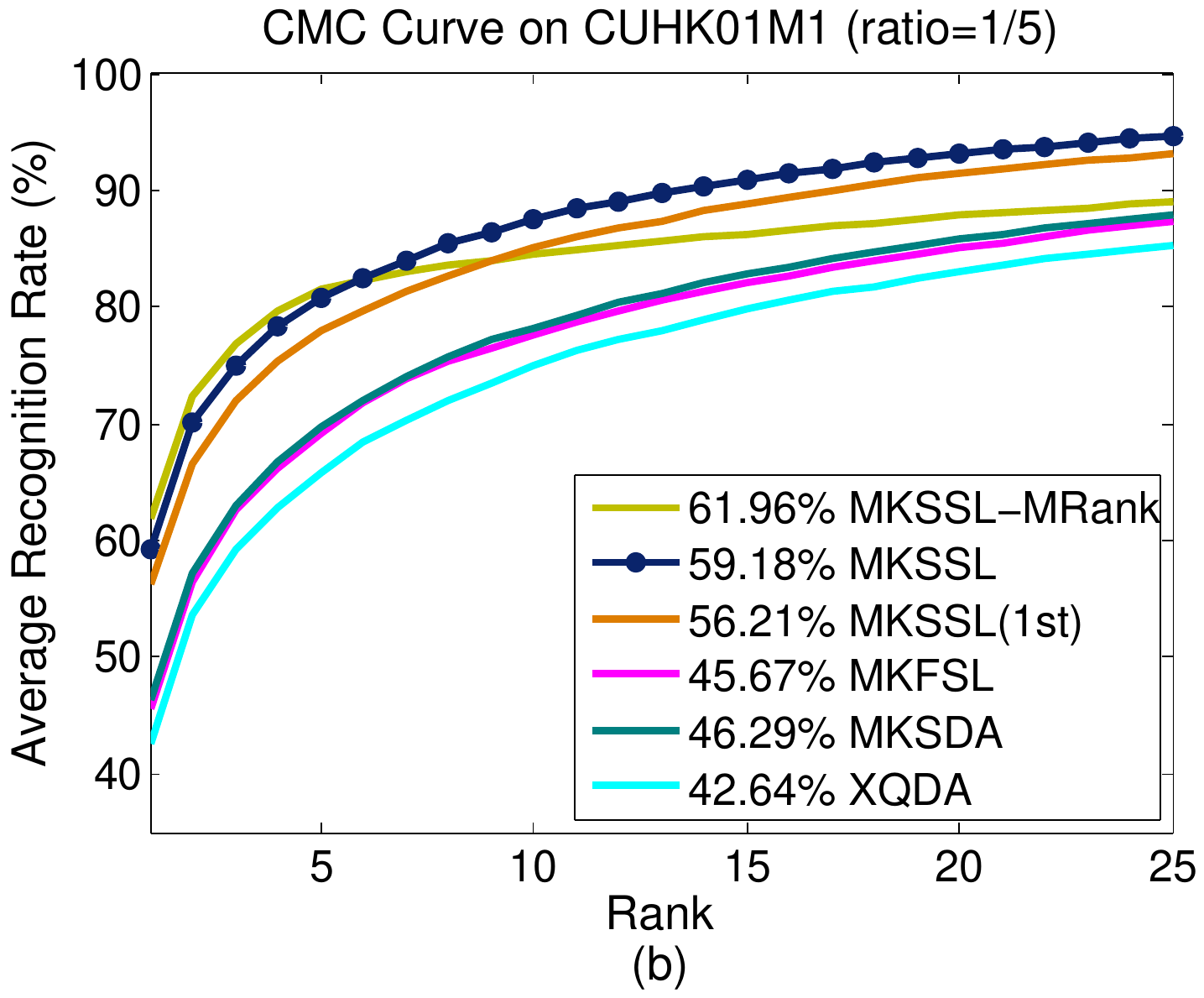} \\
 			\includegraphics[width=2.2in]{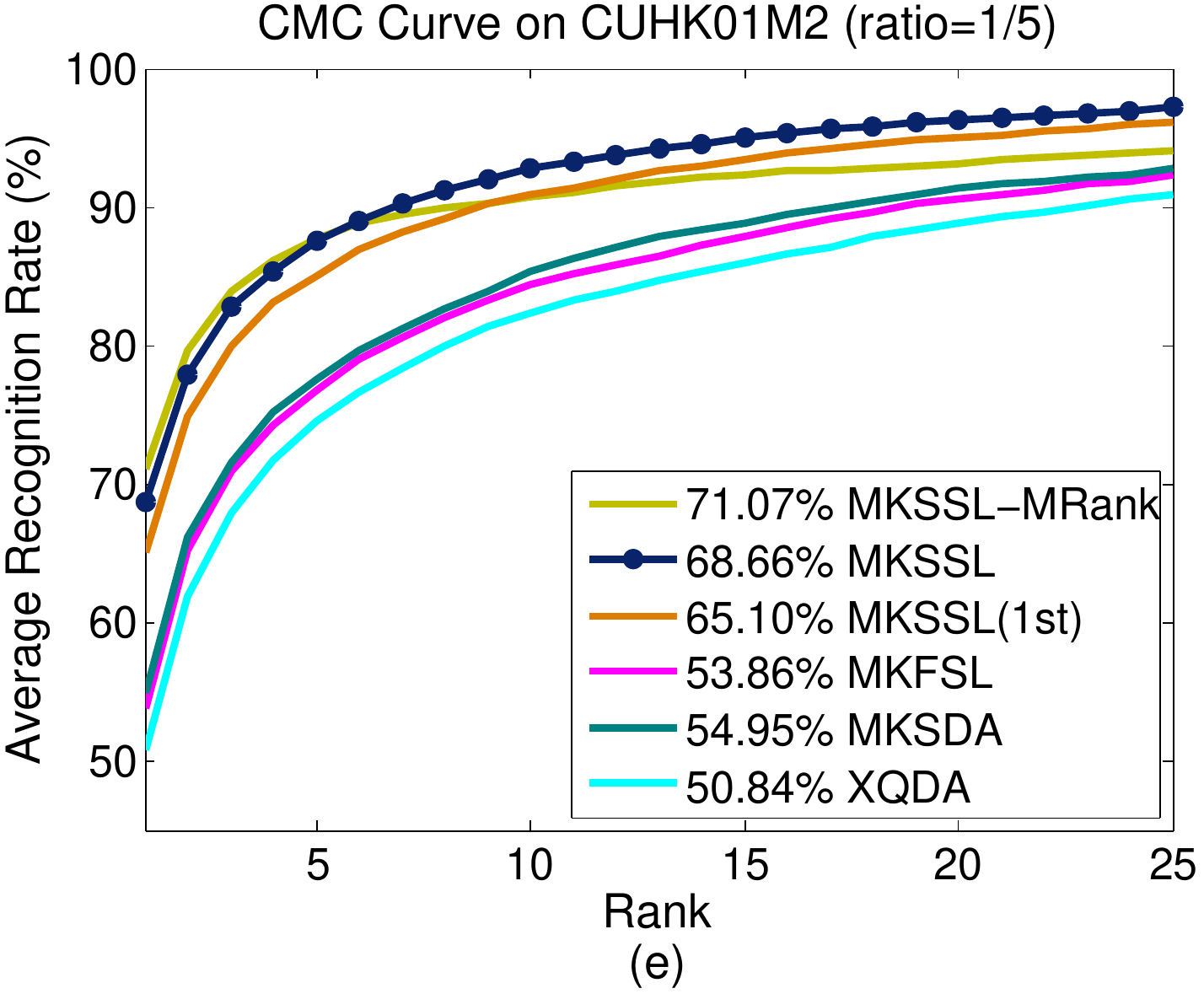} 
 		\end{minipage}
 		\begin{minipage}[b]{0.31\textwidth}
 			\includegraphics[width=2.2in]{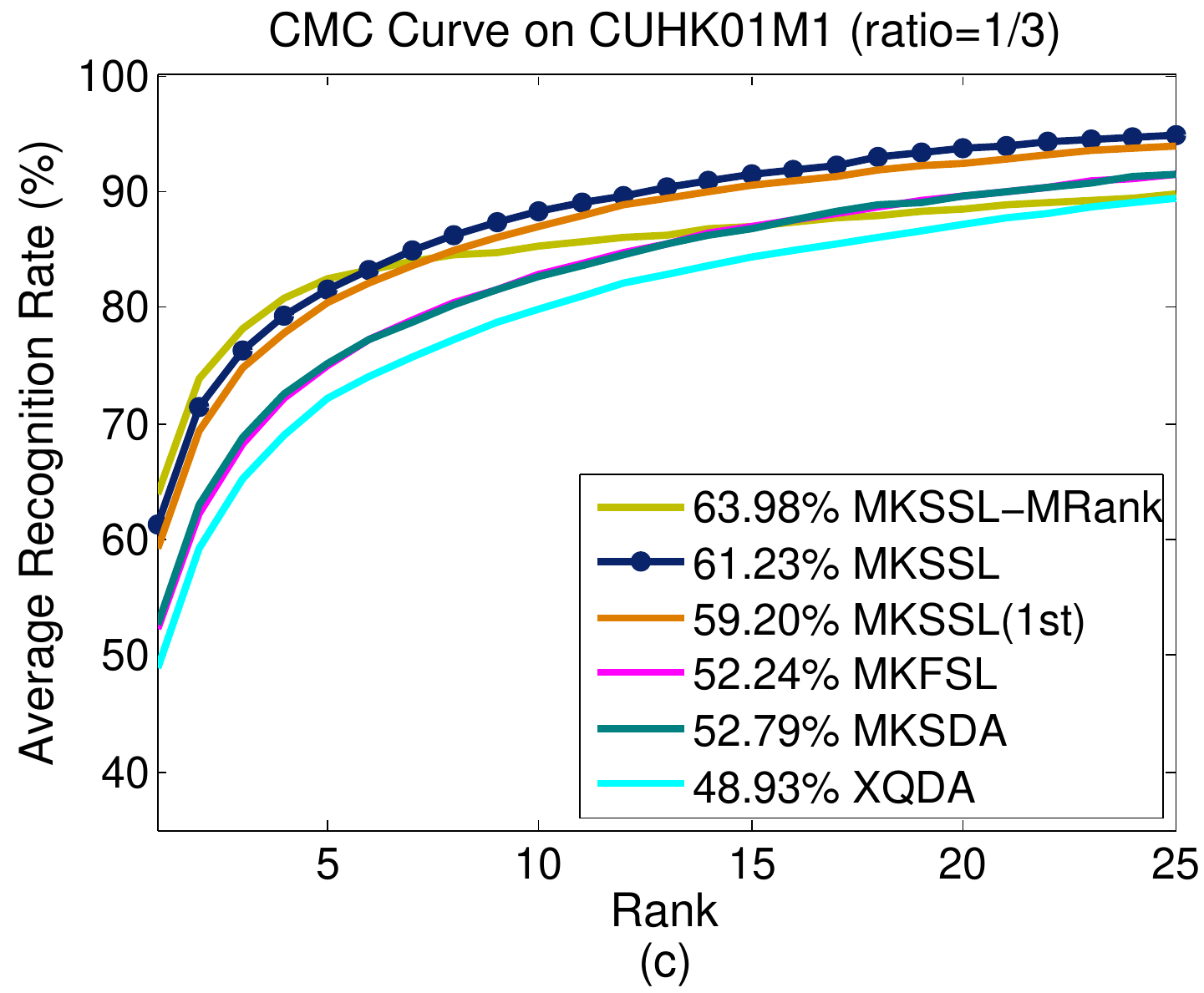} \\
 			\includegraphics[width=2.2in]{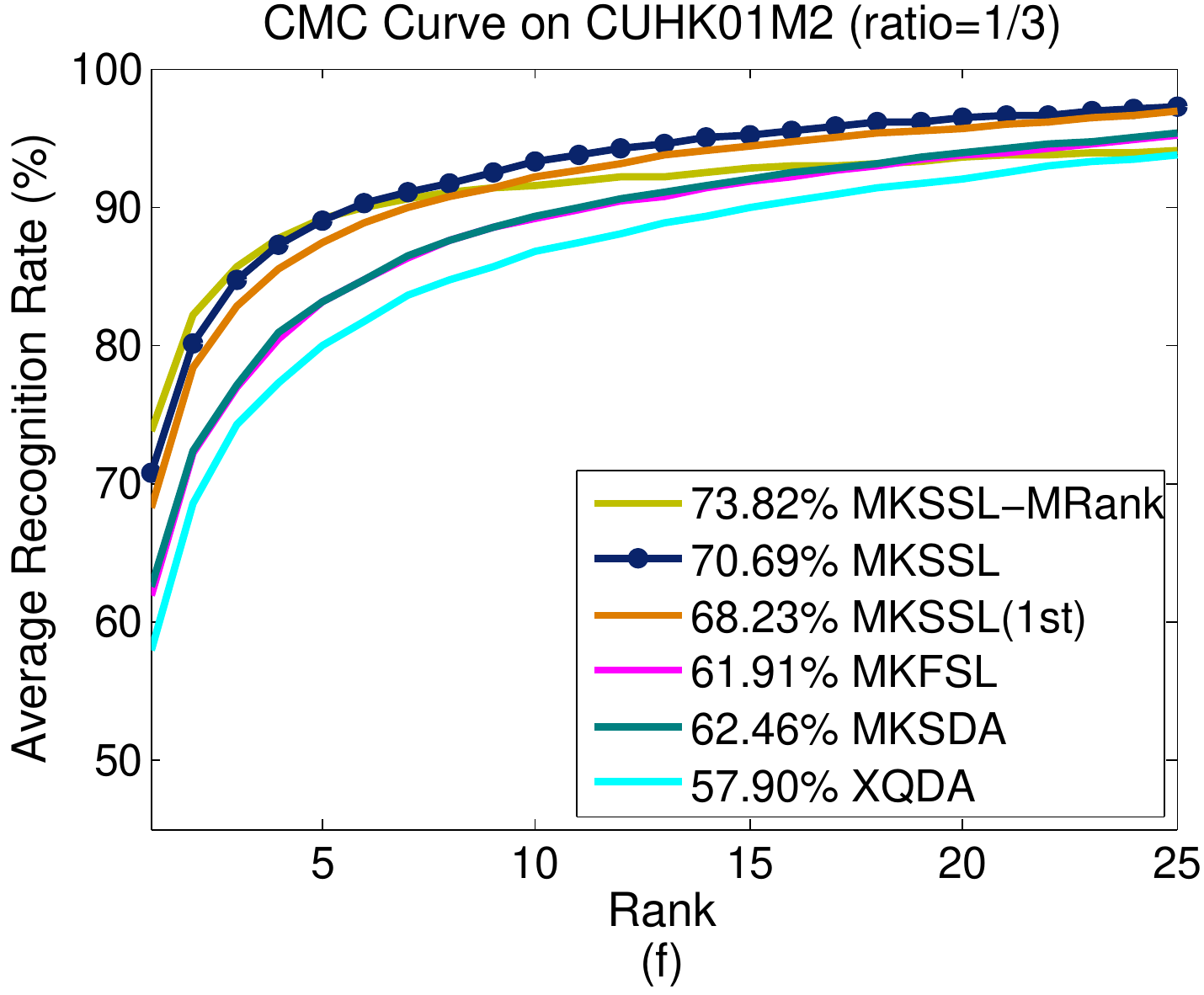} 
 		\end{minipage}
 		\vspace{-0.15in}
 		\caption{Performance comparison of our approach with different baselines on the CUHK01 dataset with different settings of $ratio$. Both the single-shot matching (M=1) and the multi-shot matching (M=2) are applied. Rank-1 recognition rates are shown in the legends.}
 		\vspace{-0.15in}
 		\label{figure4}
 	\end{figure*}
\subsection{Results of Semi-Supervised Re-ID}\label{section4.2}
In this subsection, we evaluate the performance of the proposed approach in the semi-supervised setting on the VIPeR, CUHK01, PRID2011, PRID450S, GRID, and 3DPeS datasets, compared with the baseline methods and reported results of state-of-the-art methods.

\subsubsection{Performance on the VIPeR Dataset}
The testing protocol for VIPeR is to randomly select 316 persons for training and 316 persons for testing. The training set is further split into two groups: one is labeled and the rest is unlabeled, according to the proportion $ratio$ of labeled data.
To better evaluate the performance, we conduct multiple experiments on VIPeR by setting $ratio$ to $1/7$, $1/5 $, and $1/3$, respectively. 

We show the performance comparison of the proposed MKSSL approach with baseline methods in Fig. \ref{figure3}. We observe that our approach performs very well even when very fewer training data are labeled. 
When $ratio=1/7$, MKSSL improves the performance of MKFSL by 14.78$\%$ at rank-1, which indicates that the abundant unlabeled data have been well utilized. Similar improvements 12.5$\%$ and 11.65$\%$ at rank-1 can also be observed when $ratio=1/5$ and $1/3$, respectively. With the increasing of $ratio$, the improvements drop gradually, since the size of unlabeled data is dropping. All the methods listed in Fig. \ref{figure3} yield better performance when $ratio$ is increasing. We observe that MKSSL improves the performance of MKSSL(1st) by nearly 5$\%$ at rank-1, which shows that the introduced iterative self-training strategy is, indeed, effective on VIPeR.
 It can also be observed that MKSSL(1st) performs better than MKSDA. The main difference between them is that the former is more like a semi-supervised extension of LPP (\cite{he2003locality}), while the latter is extended from LDA. The advantage of LPP on LDA has been demonstrated in \cite{he2003locality}.
By learning a discriminative subspace and a metric simultaneously, XQDA \cite{LOMOXQDA} performs better than MKFSL on VIPeR. Our MKSSL method surpasses XQDA by over $7\%$ when $ratio=1/3$ by exploiting unlabeled data. 
MKSSL reports the second-best rank-1 recognition rate 40.63$\%$ when $ratio=1/3$. Benefiting from the effectiveness of the manifold ranking method \cite{loy2013person}, MKSSL-MRank improves the rank-1 recognition rate of MKSSL by 1.3$\%$, 1.74$\%$, and 1.68$\%$, respectively, when $ratio=1/7, 1/5$ and $1/3$. It demonstrates that, as a post-ranking algorithm, this unsupervised technique can complement the proposed approach very well. It refines the initial ranking scores computed in the learned distance space of MKSSL to yield a better ranking result by exploiting the manifold structure of unlabeled gallery data. In the manifold space, a higher rank will be assigned to gallery instances situated near to the probe sample, whilst locally nearby instances are encouraged to have similar ranks.

We also compare the performance of our approach with the reported results of state-of-the-art semi-supervised or fully-supervised results on VIPeR in Table \ref{Table2}.
 From the results shown in Table \ref{Table2}, we observe that MKSSL performs better than existing semi-supervised results on VIPeR. It mainly owes to the effectiveness of both the introduced self-trained subspace learning approach and the robust GOG descriptor. To compare MKSSL with LDNS \cite{zhang2016learning}, we also provide the result of MKSSL in Table \ref{Table2} using the same LOMO descriptor. 
 We observe that LOMO+LDNS \cite{zhang2016learning} performs slightly better than LOMO+MKSSL on VIPeR. It may be because \cite{zhang2016learning} learns a higher dimensional projection by combining pseudo classes and labeled classes into a new training set, while our method keeps unlabeled data in a Laplacian regularization term and do not increase the dimension of the final subspace although we have exploited abundant unlabeled data. In comparison, our approach is able to obtain more robust performance but at the cost of a small drop in recognition rate. 
  It can also be observed in Table \ref{Table2}, our semi-supervised results are almost comparable to the reported results of most state-of-the-art fully supervised methods while using much fewer labeled data.
	  
	  \begin{table*}[tbp]
	  	\caption{Performance comparison (CMC@rank-r, $\%$) of our approach with the reported results of state-of-the-art fully-supervised methods on the CUHK01 dataset. Both the single-shot matching (M=1) results and the multi-shot matching (M=2) results are shown.
	  		A larger number indicates a better result. }{
	  		\renewcommand\arraystretch{1.2}
	  		\begin{center}
	  			\begin{tabular}{c|l|c|c|c|c|c|c}
	  				\hline				
	  				\multicolumn{2}{c|} {\multirow{2}{*}{\textbf{CUHK01}} }    &\multicolumn{3}{c|}{M=1} &\multicolumn{3}{c}{M=2}  \\\cline{3-8}
	  				\multicolumn{2}{c|}{}    & r=1 & r=5& r=20 & r=1 & r=5  & r=20\\
	  				\hline					   
	  				{Semi-supervised}& MKSSL-MRank [Ours] & 64.0 & 82.4& 88.6 & 73.8 & 89.1 & 93.5\\
	  				{ $ratio=1/3$} & MKSSL [Ours]& 61.2 & 81.6 & 93.7 & 70.7 & 89.0 & 96.4\\
	  				\hline \hline
	  				&	MKFSL [Ours]   & 62.0 & 82.9  & 94.2 & 72.2  & 89.4 & 97.0  \\\cline{2-8}
	  				&	GOG+XQDA \cite{GaussianReIDDescriptor} & 57.8& 79.1  & 92.1 & 67.3 & 86.9  & 95.9  \\
	  				& LOMO+LDNS  \cite{zhang2016learning}         &- &- &- & 65.0& 85.0 & 94.4 \\
	  				& LOMO+LSSCDL \cite{HuchuanLu_Sample-Specific_SVM_Learning}&- &-  &- & 66.0&- &- \\
	  				& LOMO+XQDA \cite{LOMOXQDA}                     & 49.2 & 75.7  & 90.8& 63.2 & 84.0  & 93.7\\
	  				{Fully-supervised}	   &        CPDL \cite{ReID_projectiveDictionary_IJCAI}    & 59.5 & 81.3 & 93.1&- &-  &-\\
	  				{ $ratio=1$} 	   & MCPCNN\cite{cheng2016person}      & {53.7}   & 91.0 & 96.3      &{-} &{-} &{-}\\  	
	  				& DeepRank\cite{DeepRankreID-TIP2016}  & 50.4  & 84.0 & 91.3  &{-} &{-} &{-}\\ 
	  				&	Deep \cite{ahmed2015improved}   & {47.5} & {80.0} & {-}    &{-} &{-} &{-}\\
	  				&       LOMO+MLAPG \cite{MLAPG}                                          &-  &-  &-   & 64.2 & 85.4  & 94.9\\
	  				&        Ensemble\cite{MetricEnsemble}       & 53.4 & 76.4  & 90.5 &- &-  &-\\
	  				& MCKCCA\cite{lisanti2016mckcca}       & 56.6 & -& 92.0 & 69.5 &- & 96.2 \\
	  				\hline		 		
	  			\end{tabular}
	  	\end{center}}
  		\vspace{-0.2in}
	  	\label{Table3}
	  \end{table*}
\subsubsection{Performance on the CUHK01 Dataset}
We randomly partition the CUHK01 dataset into 486 persons for training and 485 persons for testing. The proportion of labeled data $ratio$ is set to $1/7$, $1/5 $, and $1/3$ respectively. Correspondingly, there are 70, 98, and 162 persons involved respectively. To the best of our knowledge, we are the first to report semi-supervised results on CUHK01.

Fig. \ref{figure4} shows the performance comparison of our approach with different baseline methods. 
From the single-shot results illustrated by Fig. \ref{figure4} (a), Fig. \ref{figure4} (b), and Fig. \ref{figure4} (c), we can observe that MKSSL outperforms MKFSL by 17.03$\%$, 13.51$\%$, and 8.99$\%$ at rank-1 respectively. 
It reveals that the performance of the fully-supervised MKFSL method can be improved significantly by utilizing unlabeled data. With the increasing of unlabeled data, the contribution of unlabeled data becomes more obvious.
Besides, MKSSL improves the rank-1 recognition rates of MKSSL(1st) by 4.81$\%$, 2.97$\%$, and 2.03$\%$, respectively, when M=1. It shows that the iterative self-training strategy has enhanced the learning performance effectively.
Our approach also outperforms the classic semi-supervised MKSDA method and the fully-supervised XQDA method. MKSSL-MRank improves the rank-1 recognition rate of MKSSL by over 2$\%$ on CUHK01.
Similar improvements can also be observed in Fig. \ref{figure4} (d), Fig. \ref{figure4} (e), and Fig. \ref{figure4} (f), when performing multi-shot matching.

We also compare the performance of our approach with the reported results of state-of-the-art fully-supervised methods on CUHK01 in Table \ref{Table3}. 
From the results shown in Table \ref{Table3}, we observe that the results of our semi-supervised approach outperform the reported results of most state-of-the-art fully-supervised methods while using very fewer labeled data. 
It indicates that we may not need to require all the training data to be labeled for a large re-ID dataset. We can obtain satisfactory performance by only labeling a small proportion of training data if using an effective semi-supervised learning strategy.

\subsubsection{Performance on the PRID2011, PRID450S, GRID, and 3DPeS Datasets}
\begin{figure*}[tbp]
	\begin{center}
		\includegraphics[width=1.75in]{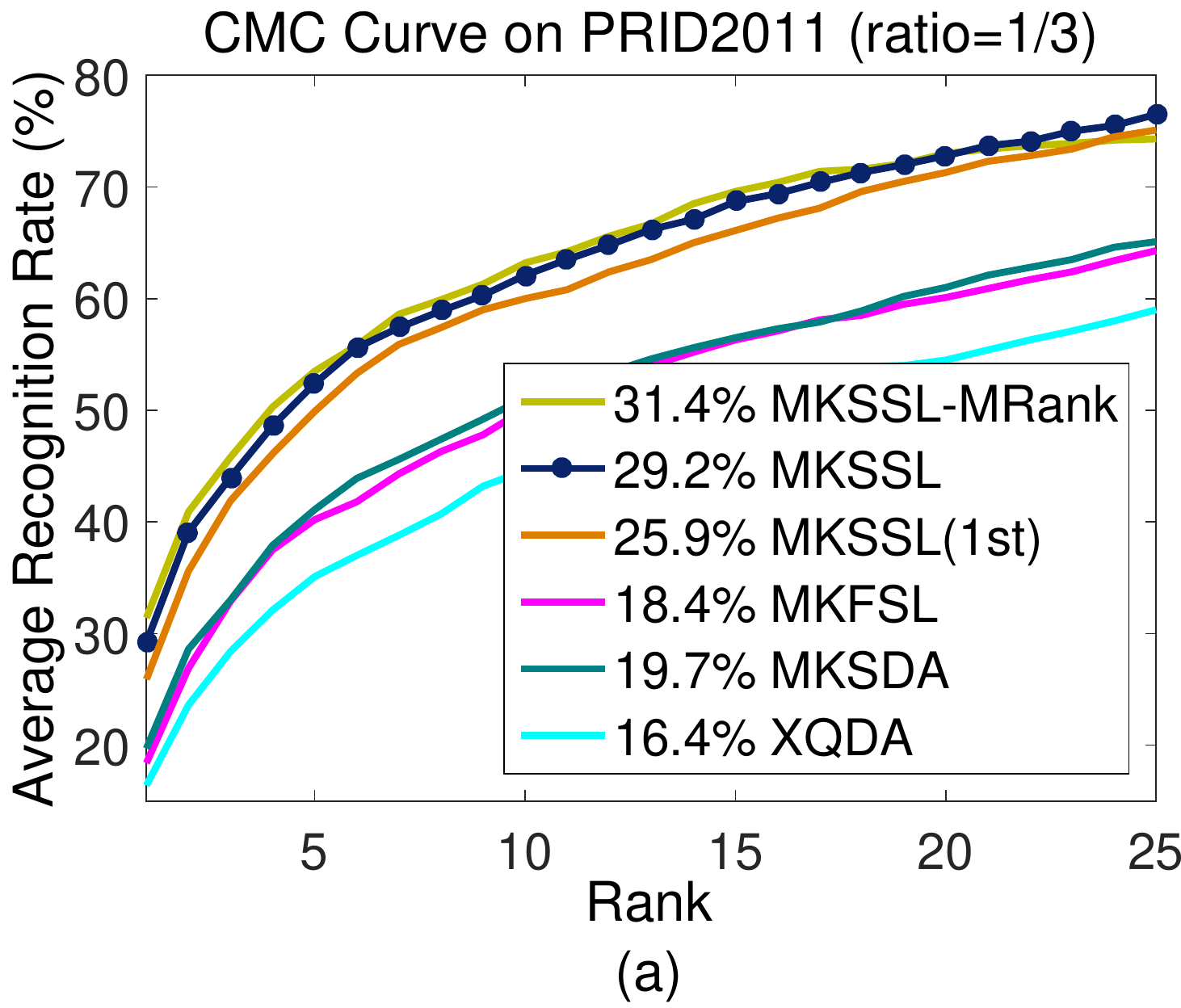} 
		\includegraphics[width=1.75in]{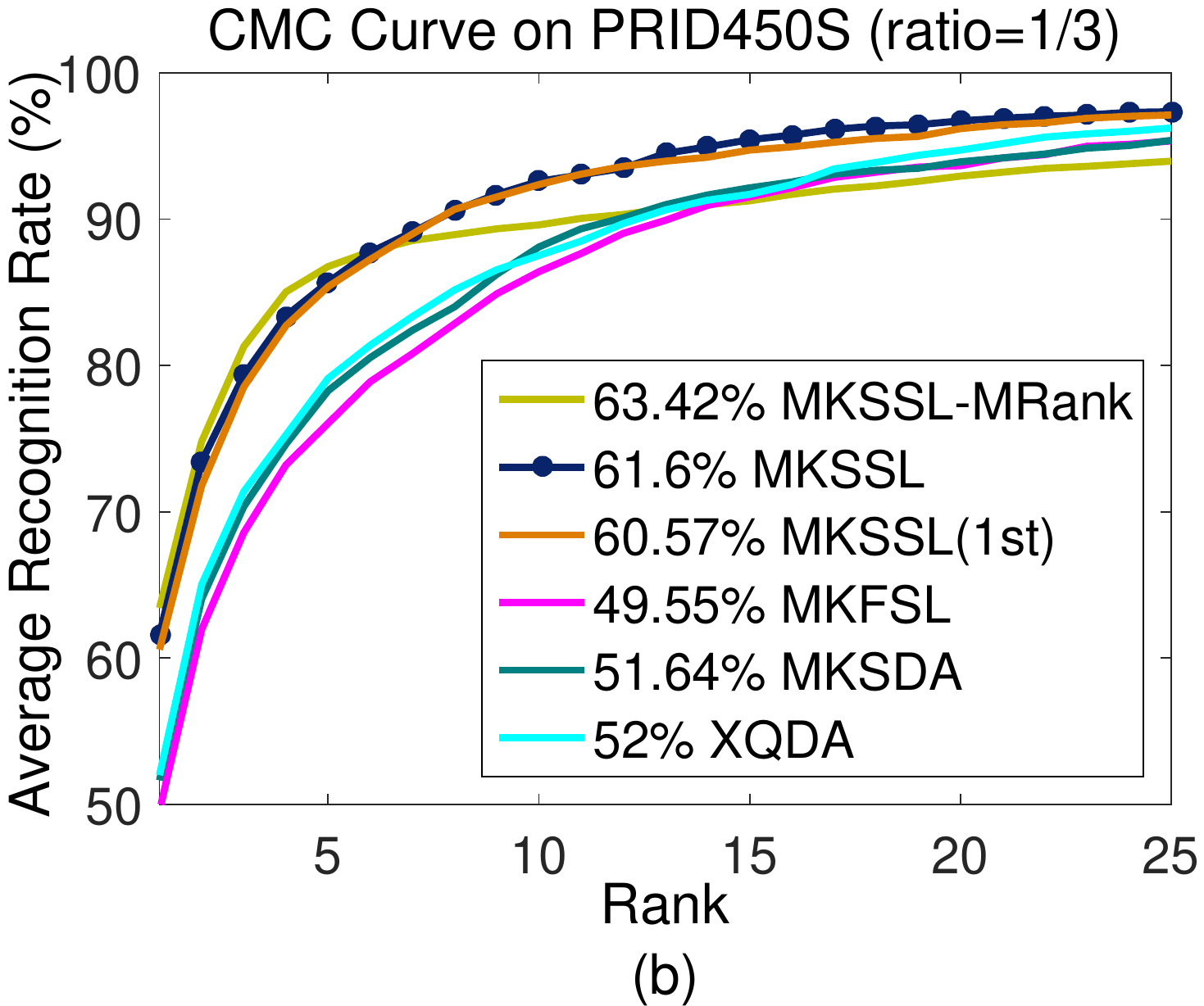} 
		\includegraphics[width=1.75in]{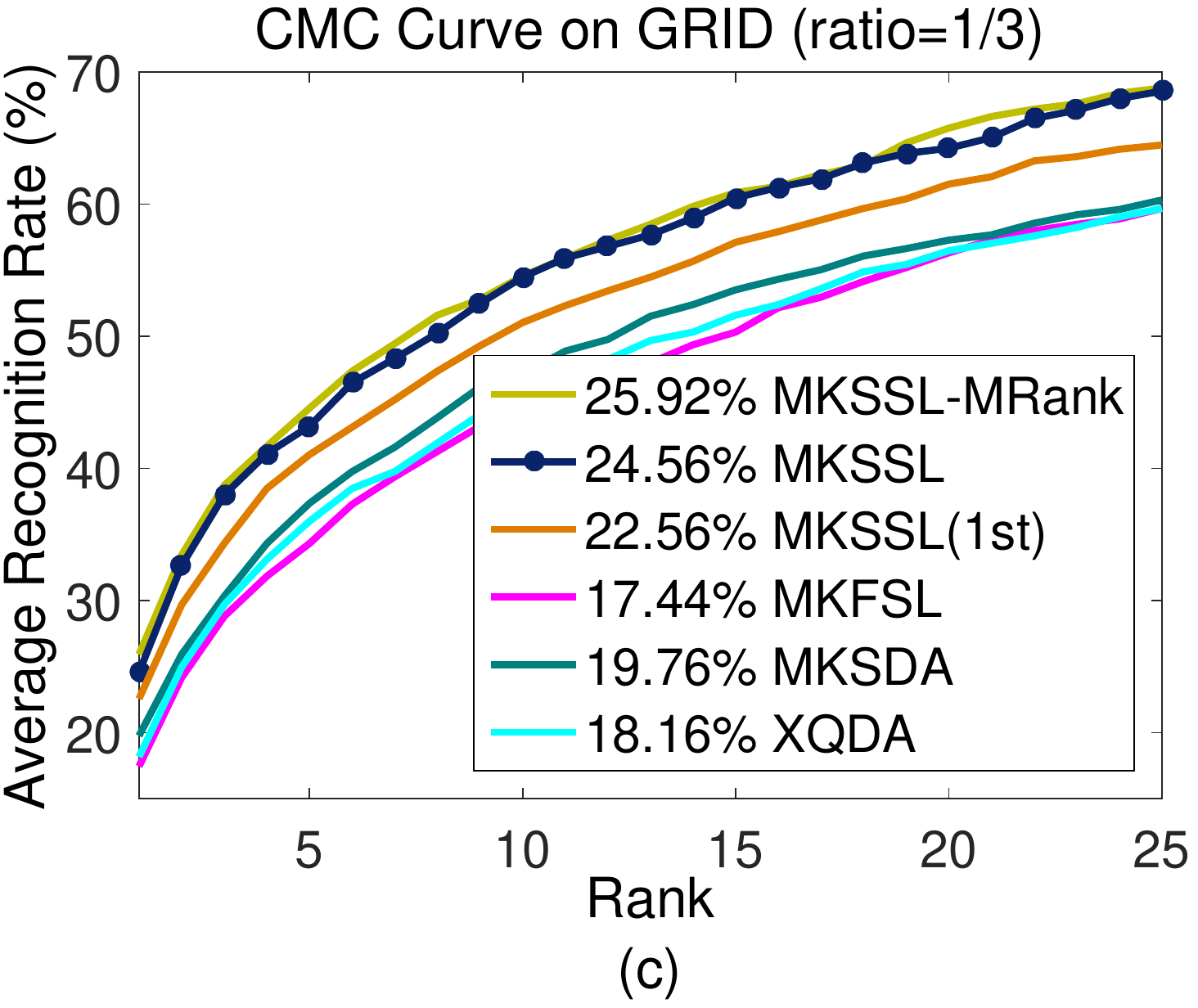} 
		\includegraphics[width=1.75in]{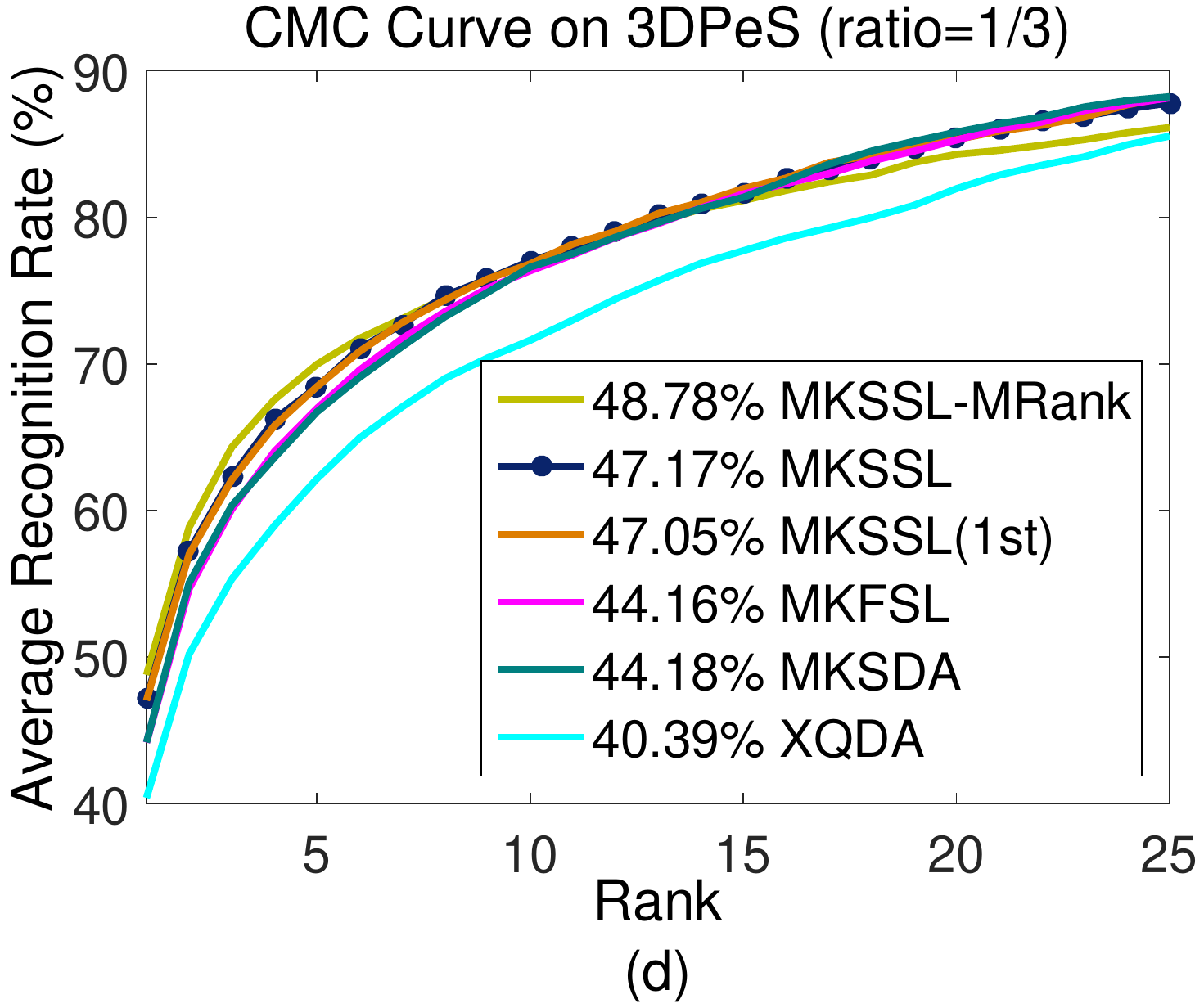} 
	\end{center}
	\vspace{-0.2in}
	\caption{Performance comparison of the proposed approach with different baselines on the six datasets: (a) PRID2011, (b) PRID450S, (c) GRID, and (d) 3DPeS. Rank-1 recognition rates ($\%$) are shown in the legends.}
	\label{figure5}
		\vspace{-0.15in}
\end{figure*}

For the evaluations on the PRID2011, PRID450S, GRID, and 3DPeS datasets, we use 100, 225, 125, and 95 individuals, respectively, for training. The remaining available individuals are used for testing. The proportion of labeled individuals in the training set is set to $ratio=1/3$ for the four datasets.
Fig. \ref{figure5} shows the CMC performance of the proposed approach and baseline methods on the four datasets with $ratio=1/3$. As shown in Fig. \ref{figure5}, by leveraging the abundant unlabeled data, our MKSSL method improves the rank-1 recognition rates of MKFSL by 10.8$\%$, 12.05$\%$, 7.12$\%$, and 3.16$\%$, respectively, on the four datasets. MKSSL improves the performances of MKSSL(1st) at rank-1 by 3.3$\%$, 1.03$\%$, 2$\%$, and 0.12$\%$, respectively. 
We observe that the iterative learning strategy of MKSSL yields nearly no improvement on MKSSL(1st) in Fig. \ref{figure5} (d). Since 3DPeS is a multi-shot dataset in which the number of images for each person varies from 2 to 26 images. Compared to other five datasets, which only have 2 or 4 images for one individual, the neighborhood structure in 3DPeS can be discovered more easily by only one iteration. 

We compare the performance of our approach with the reported results of state-of-the-art semi-supervised or fully-supervised methods on the four datasets in Table \ref{Table4}, Table \ref{Table5}, Table \ref{Table6}, and Table \ref{Table7}, respectively. We observe that, by using the GOG descriptor, our semi-supervised results can outperform the reported results of most fully-supervised methods, while using much fewer labeled data.	
  
  \begin{table}[htbp]\scriptsize
  	\caption{Performance comparison (CMC@rank-r, $\%$) of our approach with the reported results of state-of-the-art semi-supervised or fully-supervised methods on the PRID2011 dataset. A larger number indicates a better result. }{
  		\begin{center}
  			\renewcommand\arraystretch{1.2}
  			\begin{tabular}{c|l|c|c|c|c}
  				\hline
  				\multicolumn{2}{c|} {\textbf{PRID2011}}     & r=1 & r=5 & r=10 & r=20 \\
  				\hline
  				\multirow{5}{*}{ {Semi-supervised}} & MKSSL-MRank [Ours] & 31.4  & 53.5 & 63.2 & 73.0 \\
  				& MKSSL [Ours] & 29.2  & 52.4  & 62.1 & 72.8 \\
  				& LOMO+MKSSL [Ours] & 21.1   & 42.3  & 58.1 &  71.2\\
  				\cline{2-6}
  				{ \textit{ratio} = 1/3}		   & DLIterLap \cite{IterativeLaplacian} & 22.1 & 45.3 & 56.5 & 66.3\\
  				& LOMO+LDNS \cite{zhang2016learning} & 24.7 & 46.8 & 58.2 & 68.2\\
  				\hline\hline
  				\multirow{9}{*}{ {Fully-supervised}} & MKFSL [Ours] & 34.2 & 58.0 & 66.6 & 78.2\\	\cline{2-6}	
  				& GOG+XQDA \cite{GaussianReIDDescriptor}    & 35.9 & 60.1 & 68.5 & 78.1\\			  						  							
  				& LOMO+LDNS \cite{zhang2016learning} & 29.8 & 52.9 & 66.0 & 76.5\\
  				& Ensemble \cite{MetricEnsemble}       & 17.9 & 39.0 & 50.0 & 62.0\\
  				& Mahalanobis \cite{PRID_MDLReID}   & 16.0 &-& 41.0 & 51.0\\
  				{ \textit{ratio} = 1}        & DLIterLap \cite{IterativeLaplacian}     & 25.2 & 51.9 & 62.9 & 71.6\\
  				& UnL1Graph \cite{kodirov2016person}     & 30.1 & - & - & -\\				  					           
  				&  LOMO+XQDA \cite{LOMOXQDA}              & 26.7 & 49.9 & 61.9 & 73.8\\	
  				& MCKCCA\cite{lisanti2016mckcca}  & 26.7 & -& 62.1 & 73.3\\			  					          
  				\hline 			
  			\end{tabular}	
  	\end{center}}
  	\vspace{-0.2in}
  	\label{Table4}
  \end{table}

  \begin{table}[htbp]\scriptsize
  	%	\vspace{-0.1in}
  	\caption{Performance comparison (CMC@rank-r, $\%$) of our approach with the reported results of state-of-the-art semi-supervised or fully-supervised methods on the PRID450S dataset.
  		A larger number indicates a better result.}{
  		\renewcommand\arraystretch{1.2}
  		\begin{center}
  			\begin{tabular}{c|l|c|c|c|c}
  				\hline
  				\multicolumn{2}{c|} {\textbf{PRID450S}}     & r=1 & r=5 & r=10 & r=20 \\
  				\hline
  				\multirow{2}{*}{} 
  				{Semi-supervised} & MKSSL-MRank [Ours]& 63.4  & 86.8 & 89.6 & 92.9\\
  				{ \textit{ratio} = 1/3}			   & MKSSL [Ours]& 61.6  & 85.7  & 92.6 & 96.7 \\
  				\hline\hline
  				\multirow{10}{*}{} 
  				& MKFSL [Ours]& 66.9 & 89.3 & 94.2 & 97.7\\	\cline{2-6}	
  				& GOG+XQDA\cite{GaussianReIDDescriptor} & {68.4} & {88.8} & {94.5}& {97.8}\\
  				& LOMO+XQDA\cite{LOMOXQDA}  & 62.6 & 85.6 & 92.0 & 96.6 \\	
  				& LOMO+LSSCDL \cite{HuchuanLu_Sample-Specific_SVM_Learning} & 60.5 & {-} & 88.6& 93.6\\
  				{Fully-supervised}   	 & MirrorKMFA\cite{chen2015mirror}  & 55.4 & 79.3 & 87.8 & 93.9\\
  				{ \textit{ratio} = 1}	& MEDVL\cite{Metric_EmbeddedDictonaryReID}  & 45.9 & 73.0 & 82.9 & 91.1\\
  				& Transfer\cite{shi2015transferring}  & 44.9 & 71.7 & 77.5 & 86.7\\  
  				& Struct\cite{shen2015person}  & 44.4 & 71.6 & 82.2 & 89.8\\		
  				& SCNCD\cite{SalientColorName}  & 41.6 & 68.9 & 79.4 & 87.8\\	
  				& MCKCCA\cite{lisanti2016mckcca}   & 55.8 & -& 90.8 & 95.5\\	
  				\hline  			
  			\end{tabular}
  	\end{center}}
  	\vspace{-0.2in}
  	\label{Table5}
  \end{table}	
                  
\begin{table}[tbp]\scriptsize
	\caption{Performance comparison (CMC@rank-r, $\%$) of our approach with the reported results of state-of-the-art semi-supervised or fully-supervised methods on the GRID dataset.
		A larger number indicates a better result.}{
		\renewcommand\arraystretch{1.2}
		\begin{center}
			\begin{tabular}{c|l|c|c|c|c}
				\hline
				\multicolumn{2}{c|} {\textbf{GRID}}     & r=1 & r=5 & r=10 & r=20 \\
				\hline
				\multirow{2}{*}{} 
				{Semi-supervised}		  & MKSSL-MRank [Ours] & 25.7 & 44.2 & 53.8  & 65.4 \\
				{ \textit{ratio} = 1/3}		   & MKSSL [Ours] & 24.6  & 43.2  & 54.5 & 64.2 \\
				\hline\hline
				\multirow{5}{*}{ {Fully-supervised}} 
				& MKFSL [Ours] & 26.4 & 47.2 & 55.8 & 67.8\\	\cline{2-6}	
				& GOG+XQDA\cite{GaussianReIDDescriptor} & {24.7} & {47.0} & {58.4}& {69.0}\\		  						  							
				& LOMO+LSSCDL  \cite{HuchuanLu_Sample-Specific_SVM_Learning}& 22.4 &- & 51.3& 61.2\\
				{ \textit{ratio} = 1}	&  LOMO+XQDA \cite{LOMOXQDA}      & 16.6 &- & 41.8 & 52.5\\								  					          
				&  LOMO+MLAPG \cite{MLAPG}		                            & 15.6 &- & 40.5& 52.5\\	
				\hline						  			
			\end{tabular}
	\end{center}}
	\vspace{-0.15in}
	\label{Table6}
\end{table}	

\begin{table}[tbp]\scriptsize
	\caption{Performance comparison (CMC@rank-r, $\%$) of our approach with the reported results of the state-of-the-art semi-supervised or fully-supervised methods on the 3DPeS dataset.
		A larger number indicates a better result.}{
		\renewcommand\arraystretch{1.2}
		\begin{center}
			\begin{tabular}{c|l|c|c|c|c}
				\hline
				\multicolumn{2}{c|} {\textbf{3DPeS}}     & r=1 & r=5 & r=10 & r=20 \\
				\hline
				\multirow{2}{*}{} 
				{Semi-supervised}	   & MKSSL-MRank [Ours]& 48.8 & 69.9 & 76.9  & 84.3\\
				{ \textit{ratio} = 1/3}			   & MKSSL [Ours]& 47.2  & 68.4  & 77.0 & 85.4 \\		  \cline{2-6}
				& SVM+CRF \cite{karaman2014leveraging}    & 45.5 &- & - & -\\
				\hline\hline
				\multirow{3}{*}{	 {Fully-supervised}} 
				& MKFSL [Ours] & 54.4 & 77.9 & 86.7 & 94.1\\ \cline{2-6}			  						  							
				& Ensemble \cite{MetricEnsemble}       & 53.3 & - & - & -\\
				{ \textit{ratio} = 1}	          &  KLFDA  \cite{KLFDA}             & 54.0 & 77.7& 85.9 & 92.4\\
				\hline		  			
			\end{tabular}
		\end{center}
	}
	\vspace{-0.2in}
	\label{Table7}
\end{table}		      

\subsection{On the Iterative Self-training Strategy}
In this subsection we investigate the effect of the iterative self-training strategy and evaluate the robustness of our approach. We use the output of each iteration to perform person matching on the test set with different settings of $ratio$.
The VIPeR and CUHK01 datasets are used as two examples. 
We show the average rank-1 recognition rates of MKSSL at each iteration in Fig. \ref{figure7}. 

\begin{figure}[tbp]
	\begin{center}
		\includegraphics[width=2.2in]{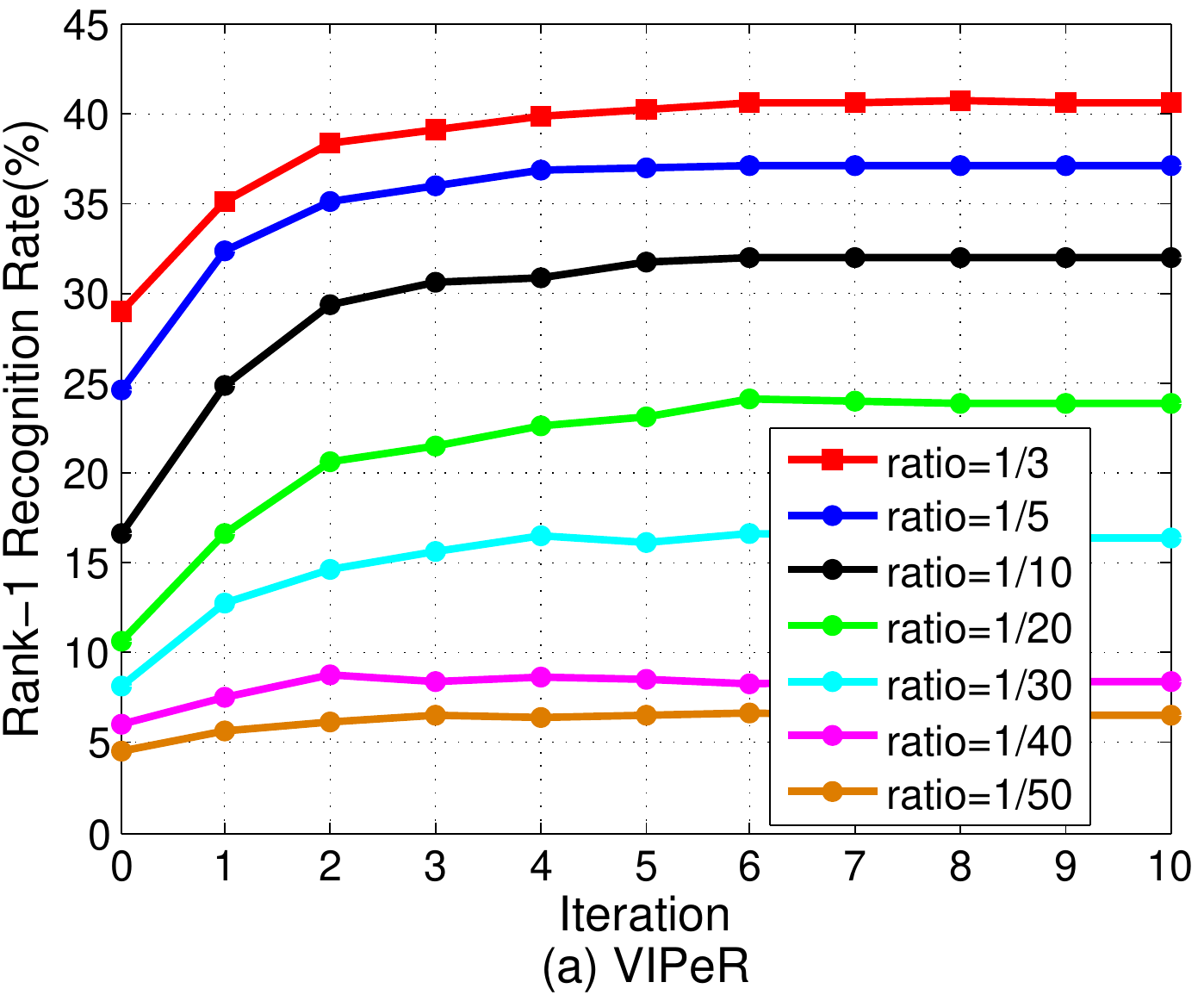} 
		\includegraphics[width=2.2in]{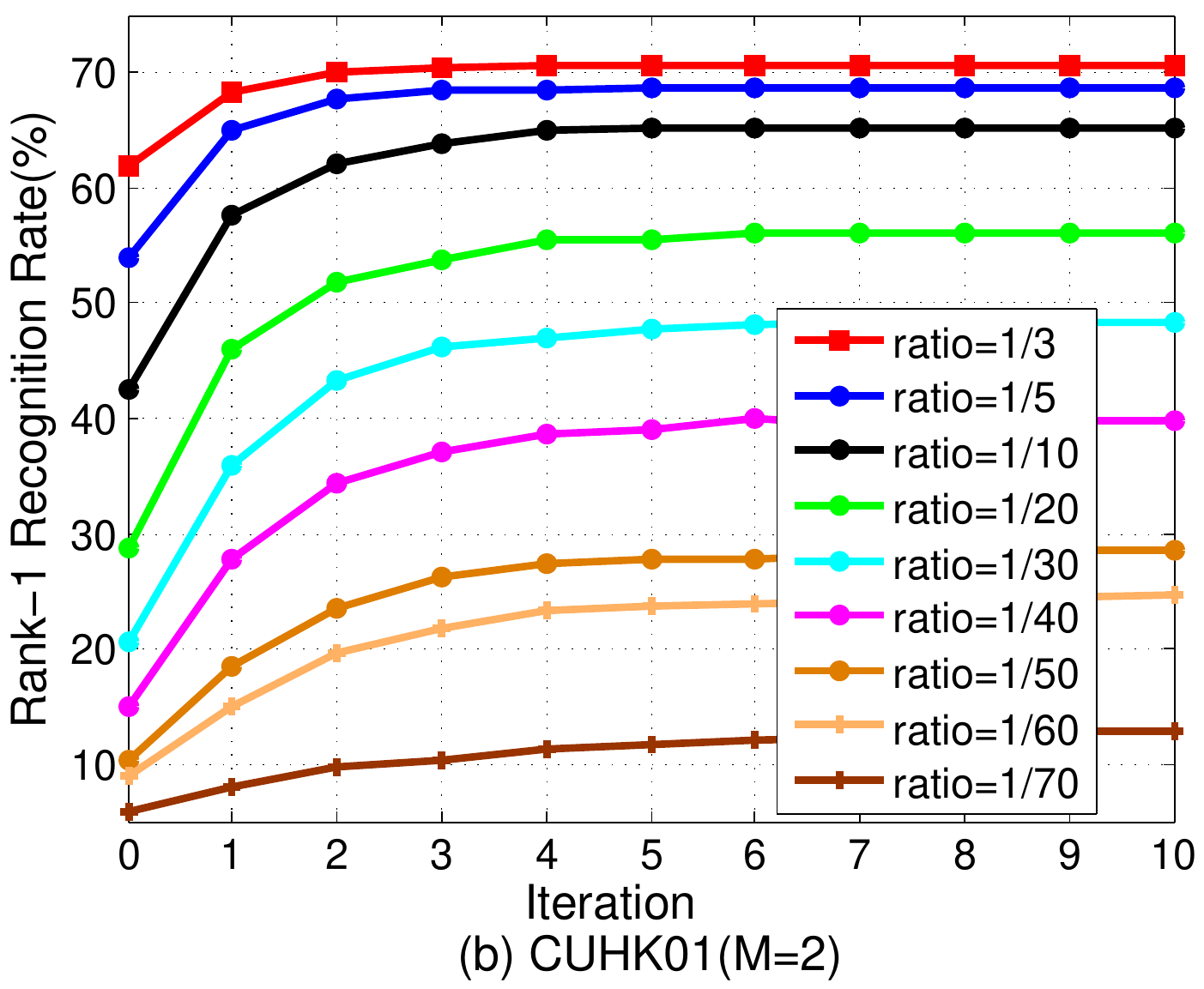} 
	\end{center}
		\vspace{-0.13in}
	\caption{Rank-1 recognition rates of MKSSL versus iteration number with different settings of $ratio$.  (a) Experiments on VIPeR; (b) Experiments on CUHK01 (M=2). Note that the rank-1 recognition rates of MKFSL are illustrated at the starting point as a comparison. }
	\label{figure6}
		\vspace{-0.2in}
\end{figure}

\begin{figure*}[tbp]
	\begin{center}
		\includegraphics[width=2.2in]{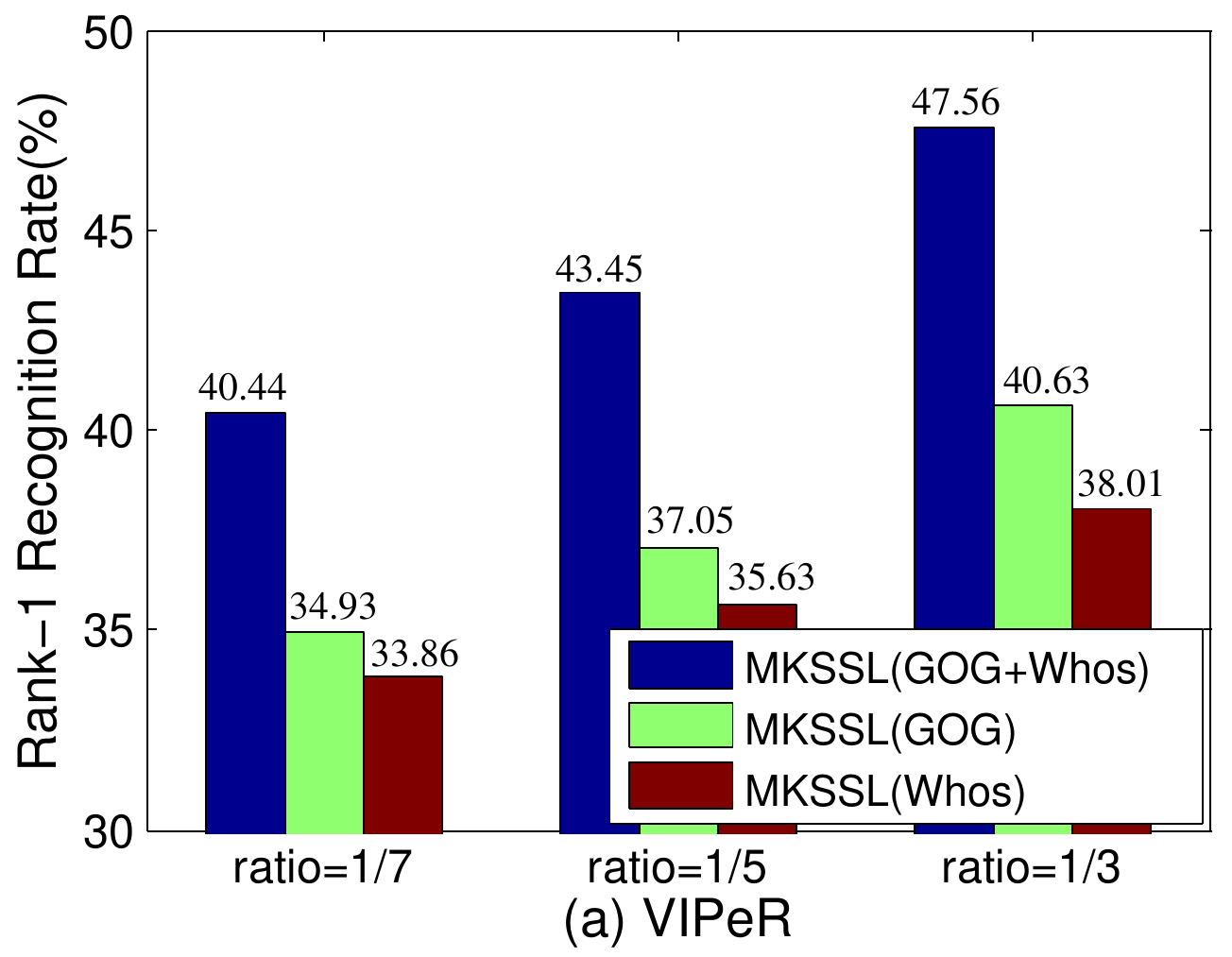} 
		\includegraphics[width=2.28in]{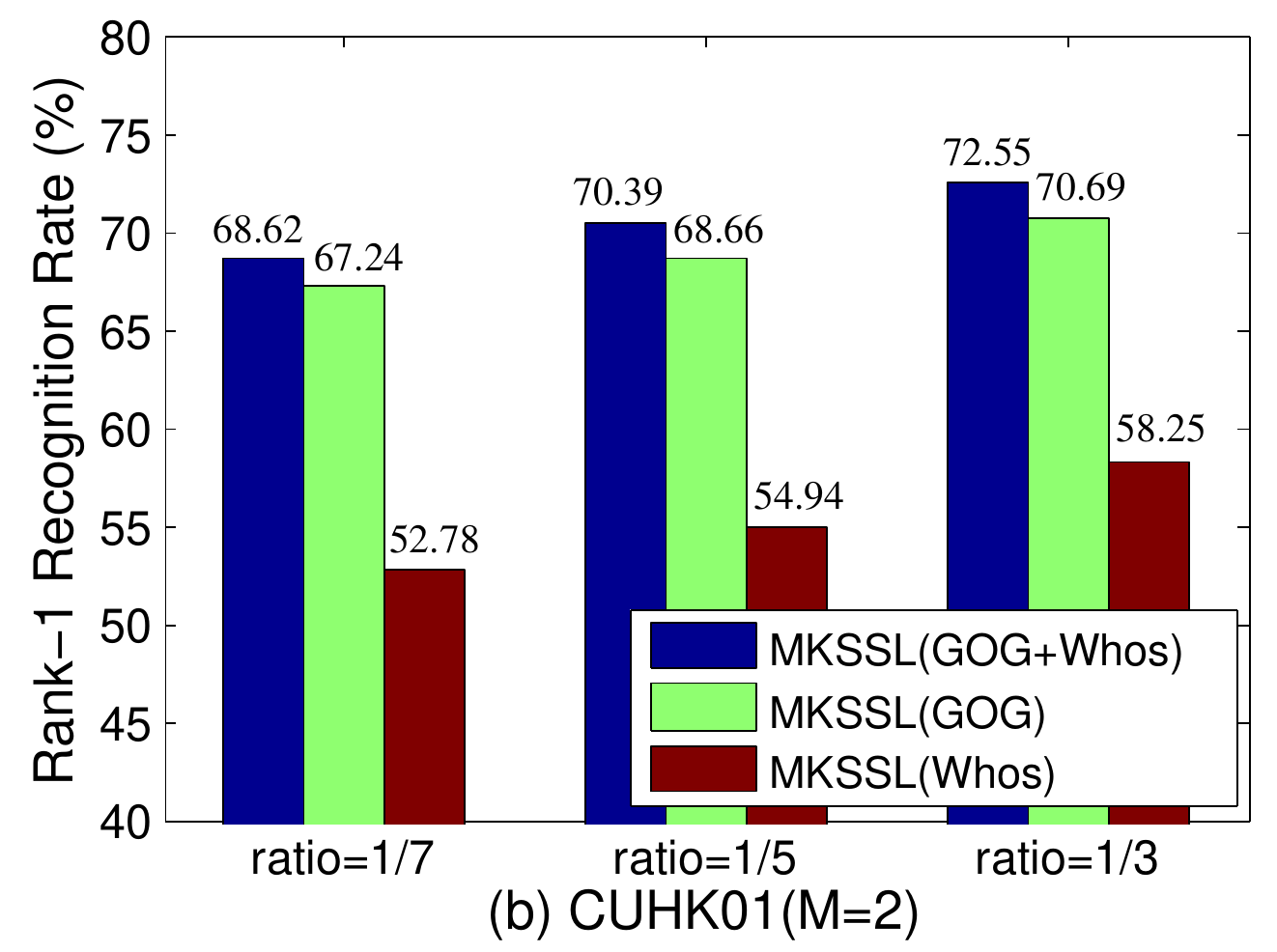} 
	\end{center}
	\vspace{-0.15in}
	\caption{The effect of multiple feature fusion with different settings of $ratio$. (a) Experiments on VIPeR; (b) Experiments on CUHK01 (M=2). }
	\label{figure7}
	\vspace{-0.15in}
\end{figure*}

First, we can observe in Fig. \ref{figure6} that the iterative self-training strategy significantly improves the performance on VIPeR and CUHK01, which demonstrates its effectiveness. A remarkable performance improvement can be observed on VIPeR when $ratio\in\{1/20, 1/10\}$ and on CUHK01 when $ratio\in\{1/40,1/30,1/20, 1/10\}$. The performance increases relatively slowly from $ratio=1/10$ to $ratio=1/3$.
With the increasing of $ratio$, the performance starts to become stable. Therefore, we can empirically conclude that very large labeled training sets may not be necessary for re-ID if applying the proposed approach in this work. When $ratio$ is very small, e.g. $ratio=1/40$ on VIPeR or $ratio=1/70$ on CUHK01, the effect of the iterative self-training strategy is not significant. Because in our setting, the subspace is initialized by the available labeled data. When $ratio=1/40$, there are only ten labeled persons on VIPeR, which is not sufficient. Therefore, $ratio$ is not suggested to be set as a very small value. The value that $ratio>=1/20$ is recommended.

Secondly, it can also be observed in Fig. \ref{figure6} that the iteration procedure converges fast in most cases. Usually it only takes three to six iterations. The larger the $ratio$ is, the faster the iteration procedure converges. The convergence time mainly depends on the size of unlabeled training data.

Besides, it should be mentioned that our approach also suffers from the model drift problem. It is a common problem in self-training based methods. Specifically, error accumulation is usually inevitable during self-training iteration procedure. This problem can be observed in Fig. \ref{figure6} (a) in which the performance degrades after 2 iterations when $ratio=1/40$. A very small $ratio$ results in a poor initial projection, which makes the model drift easily. As shown in Fig. \ref{figure6}, our approach shows empirical robustness with a large $ratio$ (e.g. 1/10) in most cases. The risk of model drift seems to have been well controlled in our approach. Because a large $ratio$ brings a good initialization, resulting in more accurate neighborhood structure. Here, the labeled term in Eq. (\ref{Eq4}) can prevent our model from going too far off a reasonable solution. By the way, to avoid the model shift, in this work we fix the dimension of the self-trained subspace as the difference of the number of labeled persons and one. We do not increase the dimension of subspace although we have exploited abundant unlabeled data, since the pseudo pairwise relationships may inevitably contain a few mismatching pairs. We may be able to obtain a better performance in a higher-dimensional self-trained subspace but at a higher risk of model drift.

\subsection{On the Complementation of Multiple Features}
In Subsection \ref{Section3.3}, we introduce the multi-kernel embedding technique \cite{shrivastava2014multiple} into the proposed approach. In the experiment results and analyses reported in Subsection \ref{section4.2}, we only utilize one feature descriptor. In this subsection, we exploit the introduced kernelization technique to consider multiple feature descriptors and evaluate the effect of this feature fusion strategy.

We use the 5138-D Whos descriptor \cite{WhosDescriptors}, together with the default 27622-D GOG descriptor used above, to learn a kernelized projection using the introduced kernelization technique. For each descriptor, 11 Gaussian kernels are constructed using the same setting in Subsection \ref{setting}. 
As shown in Fig. \ref{figure7}, we compare the performance of MKSSL using two features (MKSSL(GOG+Whos)) with that of MKSSL using only one feature (MKSSL(GOG), MKSSL(Whos)) on the VIPeR and CUHK01 datasets, respectively. We observe that, by exploiting this feature fusion strategy, MKSSL(GOG+Whos) achieves a state-of-the-art rank-1 recognition rate 47.56$\%$ on VIPeR when $ratio=1/3$, surpassing MKSSL(GOG) by nearly 7$\%$ and MKSSL(Whos) by over 9$\%$. Similar improvements are also observed when $ratio=1/5$ or $ratio=1/7$. It demonstrates that these two feature descriptors strongly complement each other on VIPeR.
 On the CUHK01 dataset, MKSSL(GOG+Whos) improves the multi-shot rank-1 recognition rate of MKSSL(GOG) by nearly 2$\%$ and that of MKSSL(Whos) by over 14$\%$, when $ratio=1/3$.
We can conclude that the introduced multi-kernel embedding strategy is flexible and effective, which can significantly enhance the performance by exploiting the complementation of multiple feature representations.

\begin{figure*}[tbp]
\begin{center}
\begin{minipage}[b]{0.32\textwidth}
 \includegraphics[width=2.1in]{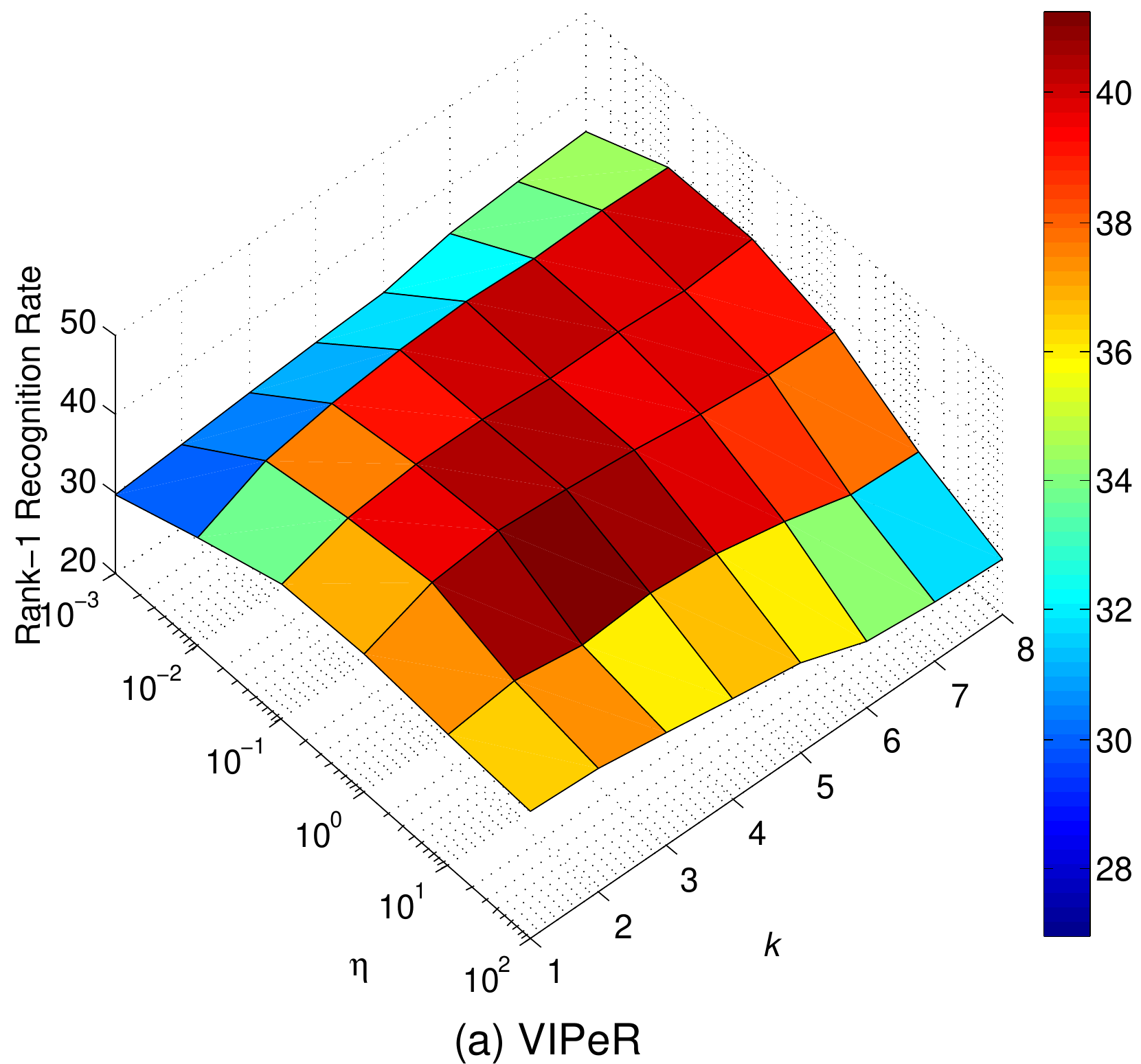} \\
 \includegraphics[width=2.1in]{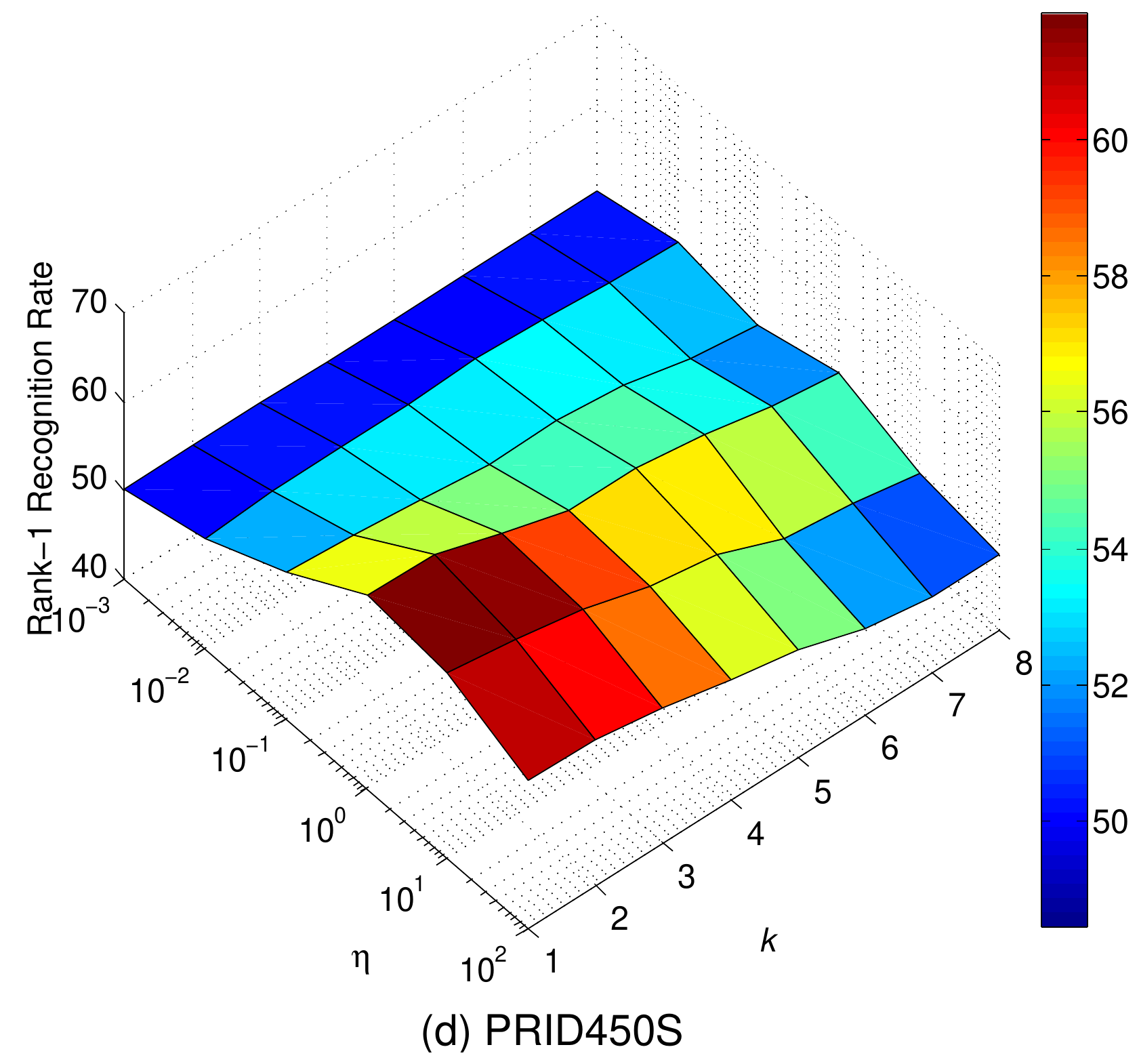} 
 \end{minipage}
 \begin{minipage}[b]{0.32\textwidth}
 \includegraphics[width=2.1in]{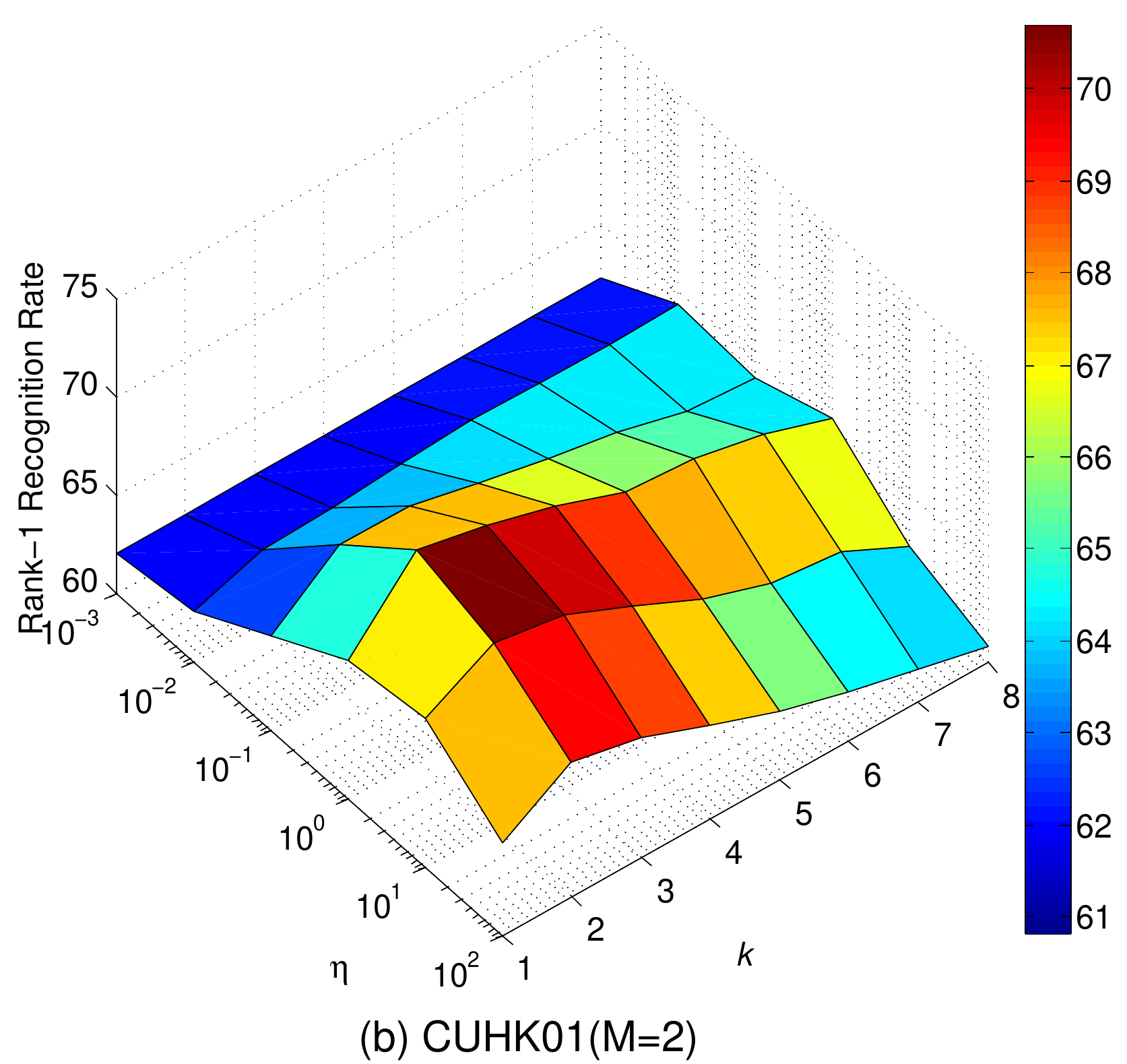} \\
 \includegraphics[width=2.1in]{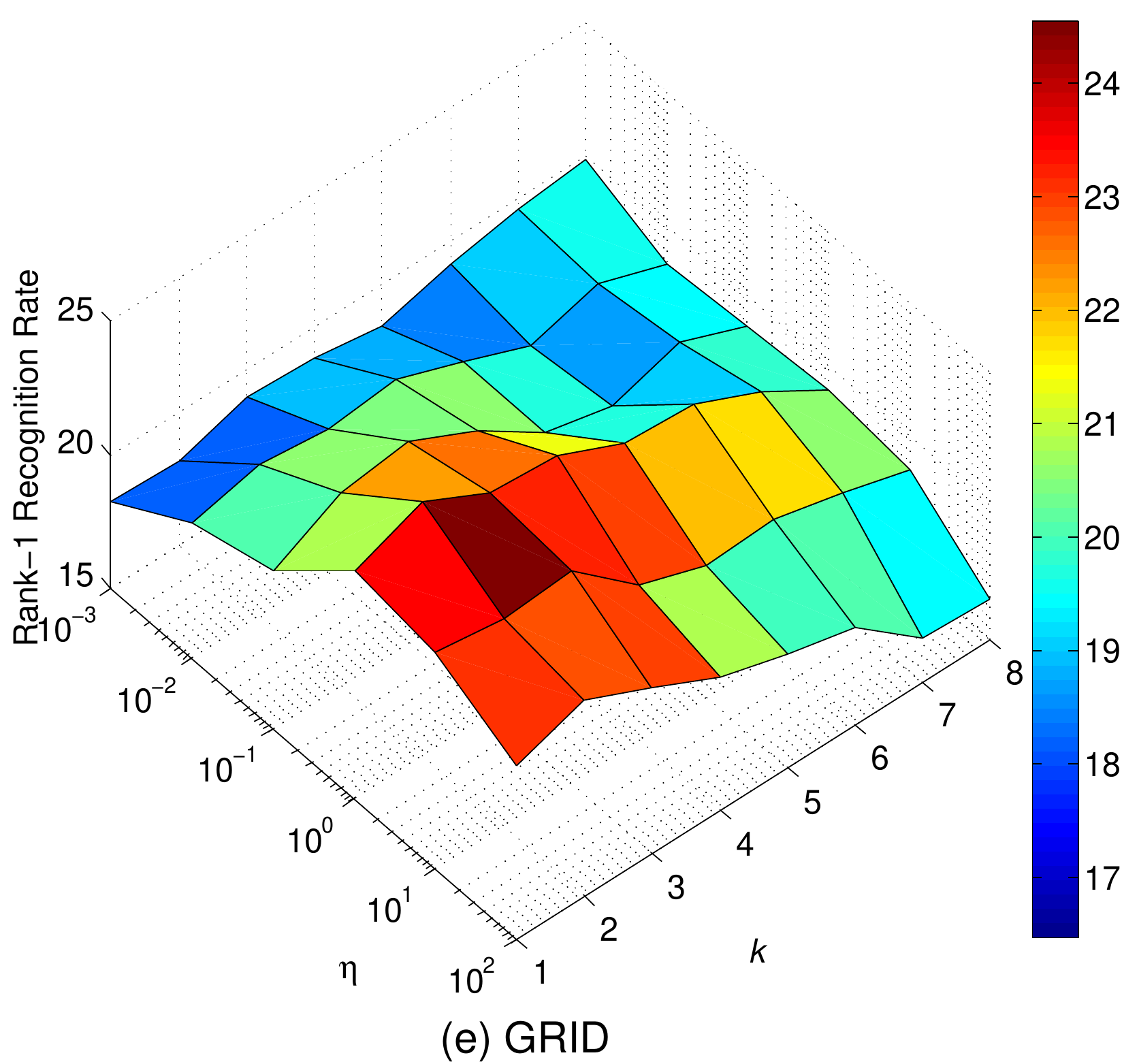} 
 \end{minipage}
 \begin{minipage}[b]{0.32\textwidth}
 \includegraphics[width=2.1in]{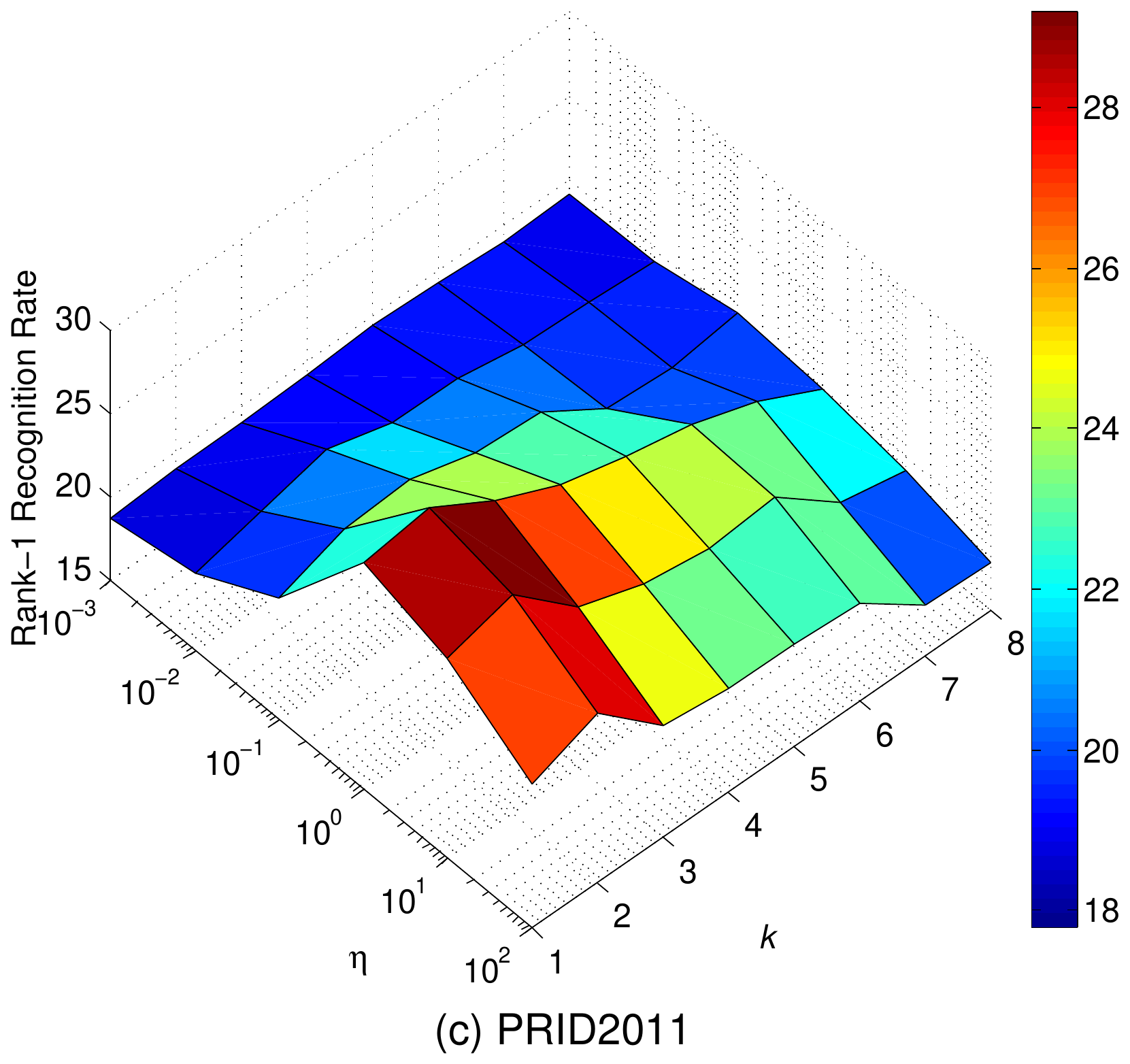} \\
 \includegraphics[width=2.1in]{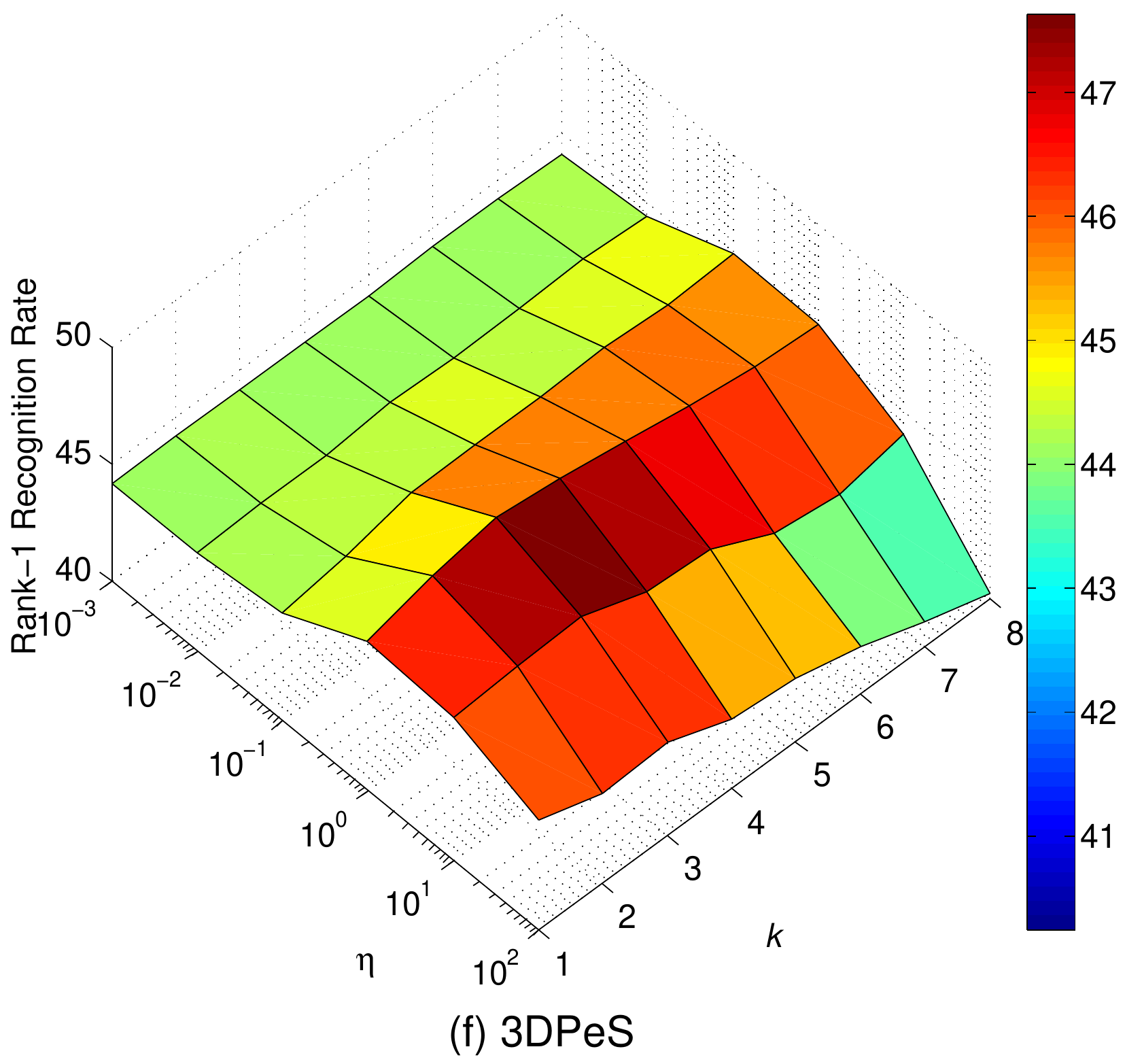} 
 \end{minipage}
\end{center} 
		\vspace{-0.15in}
 \caption{The sensitivity analyses of $\eta$ and $k$ by choosing $\eta$ from the set $\{0.001, 0.01, 0.1, 1, 10, 100\}$ and $k$ from 1 to 8 with step 1 on six datasets: (a) VIPeR, (b) CUHK01, (c) PRID2011, (d) PRID450S, (e) GRID, and (f) 3DPeS. To fairly investigate the effects of $\eta$ and $k$, we vary these two parameters simultaneously to observe the change of rank-1 recognition rate of MKSSL. Overall, our approach can perform well under a wide range of the parameter values.
 }
		\vspace{-0.15in}
 \label{figure8}
 \end{figure*}
\subsection{On the Sensitivity of Parameters}
In this subsection, we evaluate the effects of parameters used in this work. There are three tuning parameters in our approach. Parameter $\vartheta$ is a small positive parameter used to regularize the matrix at the right side of Eq. (\ref{Eq13}) to avoid the singularity of matrix. It is a commonly-used regularization technique in the eigenvalue problems. We empirically set $\vartheta$ to 0.01. Here, we mainly analyze the effects of the two main parameters: $\eta$ and $k$.  
 
Parameter $\eta$ in Eq. (\ref{Eq4}), Eq. (\ref{Eq5}), and Eq. (\ref{Eq13}) modulates the effects of the regularization term $\mathcal{R}(\mathbf{X}_u, \mathbf{U}, \mathbf{W}^u)$ constructed using unlabeled data. The number of nearest neighbors $k$ is vital in the kNN graph. In this work, we set them as $\eta=1$ and $k=2$ by cross-validation on the training set of CUHK01 and fix them for all datasets.
To fairly investigate the effects of $\eta$ and $k$, we observe the change of rank-1 recognition rate of MKSSL on the six datasets by varying $\eta$ and $k$ simultaneously. Specifically, the value of $\eta$ is chosen from the set $\{0.001, 0.01, 0.1, 1, 10, 100\}$, and $k$ is increased from 1 to 8 with step 1. The experimental results are illustrated in Fig. \ref{figure8}.

As observed in Fig. \ref{figure8}, our approach performs well on most datasets when $0.1\leq\eta\leq10$ and $k\in\{2,3,4\}$. On the PRID2011 and PRID450S datasets, our approach yields a good performance when $k=1$. On the 3DPeS dataset, a high rank-1 recognition rate can be observed when $k=4$, since it is a multi-shot dataset. Overall speaking, our approach can obtain good performance under a wide range of the parameter values.

\subsection{Time Complexity Analysis}
The main computational cost of our MKSSL method is dominated by two parts. 
The first part comes from the computation of the initial projection learned using labeled data only. Its complexity is stemmed from solving a kernelized eigenvalue problem, which is approximated by $O\left(n^3+rn^2\right)$, where $n$ is the number of labeled training images, which equals to the size of the kernel matrix constructed using labeled data, and $r$ is the dimension of the initial low-dimensional subspace. 
The second part is solving the kernelized eigenvalue problem in Eq. (\ref{Eq13}) using both labeled and unlabeled data. This procedure is repeated several times by applying the iterative self-training strategy.
Its complexity is approximated by $O\left(T\left((n+u)^3+r'(n+u)^2\right)\right)$, where $r'$ is the dimension of the final subspace, $u$ is the number of unlabeled training images, and $T$ is the iteration number. 
Therefore, the total complexity is $O\Big(n^3+rn^2+T\left((n+u)^3+r'(n+u)^2\right)\Big)$, which is mainly determined by the number of training images and the iteration number. The proposed MKSSL approach is implemented in Matlab on a 2.9GHz CPU PC with 32G RAM. 

Here, we present the practical runtime of MKSSL on the whole VIPeR dataset and the whole CUHK01 dataset with $ratio=1/3$. The average training and testing time on VIPeR are 2.88s and 0.02s respectively. The average iteration number on VIPeR is $5.8$. The average training and testing time on CUHK01  (M=2) are 28.06s and 0.09s respectively. The average iteration number on CUHK01 (M=2) is $4.1$.

\section{Conclusion}
In this work, we propose an effective semi-supervised approach for person re-ID which can leverage both labeled and unlabeled data. It formulates re-ID as a subspace learning problem by learning a discriminative projection to map the person images from disjoint camera views into a common subspace where person matching can be easily performed. It presents a self-training based subspace learning strategy in which the unlabeled person images are exploited by constructing the pseudo pairwise relationships. An iterative learning strategy is introduced to refine the pseudo pairwise relationships, which significantly enhances the learning performance.
It is also able to explore the complementary characteristic of multiple feature representations for re-ID. Experimental results on multiple challenging datasets have demonstrated the effectiveness and robustness of the proposed approach.

\bibliographystyle{./IEEEtran}
\bibliography{mybib}
\end{document}